\documentclass[10pt,twocolumn,letterpaper]{article}

\usepackage{cvpr_main}              %

\definecolor{cvprblue}{rgb}{0.21,0.49,0.74}
\usepackage[pagebackref,breaklinks,colorlinks,allcolors=cvprblue]{hyperref}

\usepackage{ulem}           %

\usepackage{graphicx}
\usepackage{amssymb, pifont}
\usepackage[utf8]{inputenc} %
\usepackage[T1]{fontenc}    %
\usepackage{url}            %
\usepackage{booktabs}       %
\usepackage{amsfonts}       %
\usepackage{nicefrac}       %
\usepackage{microtype}      %
\usepackage{xcolor}         %
\usepackage[english, status=draft, margin=false, inline=true]{fixme}
\fxusetheme{color}
\FXRegisterAuthor{gl}{ghl}{\color{blue}GL}
\usepackage{graphicx}
\usepackage[utf8]{inputenc} %
\usepackage[T1]{fontenc}    %
\usepackage{hyperref}       %
\usepackage{amsfonts}       %
\usepackage{nicefrac}       %
\usepackage{microtype}      %

\usepackage{tabularray}
\usepackage{colortbl}
\usepackage{arydshln}
\usepackage{minted}
\usepackage{listings}
\usepackage{xcolor}

\definecolor{mygray}{gray}{.9}
\definecolor{mygreen}{rgb}{0, 0.6, 0}
\hypersetup{
    colorlinks=true,
    linkcolor=red,
    anchorcolor=red,
    citecolor=red,
    urlcolor=red
}

\usepackage{algorithm}
\usepackage{algorithmic}
\usepackage{booktabs}
\usepackage{times}
\usepackage{epsfig}
\usepackage{amsmath}
\usepackage{amssymb}
\usepackage{caption}
\usepackage{subcaption}
\usepackage{multirow}
\usepackage[flushleft]{threeparttable}
\usepackage{pifont}%
\newcommand{\xmark}{\ding{55}}%
\usepackage{tikz}
\usepackage{tabularx}

\usepackage{color}
\usepackage{array}
\usepackage{makecell}
\definecolor{darkgreen}{RGB}{0,100,0}

\newlength\savewidth
\newcommand{\bg}{\mathbf{g}}

\definecolor{mygray}{gray}{0.9}
\definecolor{iccvblue}{rgb}{0.21,0.49,0.74}
\definecolor{mypink}{RGB}{255, 102, 178}  %

\def\paperID{} %
\def\confName{CVPR}
\def\confYear{2026}

\title{
Sky2Ground: A Benchmark for Site Modeling under Varying Altitude
}

\vspace{-2em}
\author{
Zengyan Wang, Sirshapan Mitra, Rajat Modi, Grace Lim, Yogesh Rawat \\
$\{zengyan.wang, sirshapan.mitra, yogesh\}@ucf.edu$\\
Center for Research In Computer Vision\\ University of Central Florida \\
}

\begin{document}
\maketitle

\begin{abstract}

We introduce Sky2Ground, a three-view dataset designed for varying altitude camera localization, correspondence learning, and reconstruction.
The dataset combines structured synthetic imagery with real, in-the-wild images, providing both controlled multi-view geometry and realistic scene noise.
Each of the 51 sites contains thousands of satellite, aerial, and ground images spanning wide altitude ranges and nearly orthogonal viewing angles, enabling rigorous evaluation across global-to-local contexts.
We benchmark state of the art pose estimation models, including MASt3R, DUSt3R, Map Anything, and VGGT, and observe that the use of satellite imagery often degrades performance, highlighting the challenges under large altitude variations.
We also examine reconstruction methods, highlighting the challenges introduced by sparse geometric overlap, varying perspectives, and the use of real imagery, which often introduces noise and reduces rendering quality.
To address some of these challenges, we propose SkyNet, a model which enhances cross-view consistency when incorporating satellite imagery with a curriculum-based training strategy to progressively incorporate more satellite views. SkyNet significantly strengthens multi-view alignment and outperforms existing methods by $9.6\%$ on RRA@5 and $18.1\%$ on RTA@5 in terms of absolute performance.
Sky2Ground and SkyNet together establish a comprehensive testbed and baseline for advancing large-scale, multi-altitude 3D perception and generalizable camera localization. Code and models will be released publicly for future research. Project page: \url{https://sky2ground2026.github.io/sky2ground/}
\end{abstract}

\begin{figure*}[t]
  \centering
  \vspace{-2em}
  \includegraphics[width=1\linewidth]{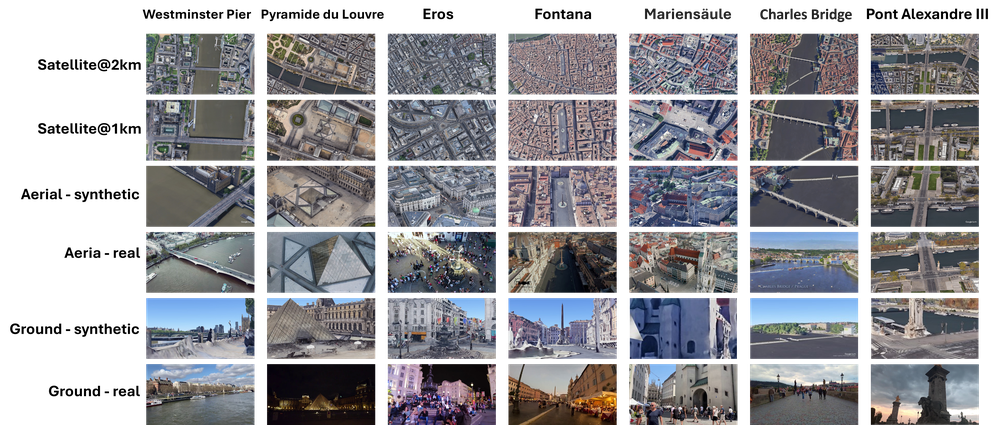}
 \caption{\textbf{Cross-view examples from the Sky2Ground dataset.} Satellite, aerial, and ground-level images for a variety of urban scenes in Sky2Ground, where each column corresponds to a unique site. 
 These examples highlight strong viewpoint and appearance variations across modalities, revealing the challenges of cross-view matching and multi-scale scene understanding. 
 Real images additionally introduce diverse lighting conditions, weather effects, and natural scene noise, further emphasizing the complexity of real-world cross-view perception.}
  \label{fig:data}
  \vspace{-1em}
\end{figure*}

\vspace{-10pt}
\section{Introduction}
\label{sec:intro}

\indent  One of the \textit{fundamental} tasks in computer vision has been to \textit{jointly} localize multi-view images and construct \textit{reliable} 3D models of a scene. This has found many commercial applications, such as geo-registering ground-images~\citep{mithun2023cross,pittaluga2019revealing,workman2015wide}, \textit{outdoor-scene} reconstruction~\citep{patel2020visual,rezaei2024sat2map,argota2025urban},  urban-view synthesis\cite{xiangli2022bungeenerf,xu2023vr,li2023matrixcity,lehtola2022digital}, and complex navigation~\citep{mithun2023cross,li2024learning,peldon2024navigating}. Classical pipelines like COLMAP~\cite{schoenberger2016mvs,schoenberger2016vote} have leveraged hand-designed features and matching heuristics, which makes them \textit{computationally-intensive} in practice. Recently, neural nets have achieved great success in inferring geometry parameters like camera poses/depth maps, thereby enabling \textit{faster} predictions\cite{mast3r_arxiv24,dust3r_cvpr24,wang2025vggt}. This progress has been largely facilitated by \textit{pre-training} on ground-aerial datasets\cite{vuong2025aerialmegadepth}. 
 However, modeling large-scale scenes across \textit{multiple} altitudes and viewpoints remains underexplored, notably \textit{within the domain of satellite imagery.}

\textit{Unlike} ground/aerial views, satellite views provide globally consistent coverage, stable geospatial reference frames, and fine spatial resolution~\citep{mari2022sat}.
Recent work uses satellite imagery for localization from oblique or UAV perspectives in GPS-denied environments~\citep{patel2020visual,mari2022sat,conte2008integrated}. We notice a \textit{lack} of datasets which provide \textit{simultaneous} access to \textit{ground/aerial/satellite} views. Furthermore, to the best of our knowledge, there is \textit{no work} focusing on joint-localization of cameras across \textit{all three-views}, i.e. ground/aerial/satellite. However, we believe this problem to be \textit{very-timely}, given the recent-efforts trying to achieve the \textit{ambitious-goal} of \textit{planet-wide} reconstruction\cite{vuong2025aerialmegadepth}.

Motivated by this, we propose a multi-altitude, cross-view dataset called \textbf{Sky2Ground} that \textit{unifies synthetic} satellite, aerial, and ground-level imagery  \textit{with real} aerial and ground imagery collected from-the-wild. 
The \textit{synthetic component} provides dense, globally consistent camera poses and structured 3D geometry, while the \textit{real component} introduces photometric variability, occlusions, and environmental complexity.
As a result, our dataset enables a \textit{thorough evaluation} of cross-view localization and 3D reconstruction pipelines. 

Furthermore, we perform a detailed benchmark study which reveals that existing pose estimation/reconstruction methods \textit{struggle under} large distributional shifts introduced by satellite/real-images. We further show that while aerial and ground modalities \textit{reinforce} each other during training, even a \textit{single} satellite image can \textit{destabilize} training, thereby highlighting the \textit{challenge of } effective multi-altitude integration in current 3D models. 

Inspired by these observations, we introduce \textit{SkyNet}, a model designed to strengthen 3D reconstruction and localization models \textit{in presence} of satellite imagery. 
SkyNet is a two-stream architecture consisting of two-components 1) a \uline{G}round-\uline{A}erial-\uline{S}atellite (GAS-encoder) which processes all modalities together. It consists of  restricted-global-attention module called \uline{M}asked-\uline{S}atellite-\uline{A}ttention (MSA) that \textit{prevents} aerial/ground modalities from interacting with satellite. 2) a Sat-Encoder which explicitly processes satellites. Furthermore, SkyNet is trained via novel curriculum-based strategies:  1) gradually sampling views which are further from each other 2) using `aerial-images' as a bridge between ground/satellite modalities. We find that these innovations enable \textit{adapting} and even improving upon a model like VGGT, thereby being able to retain advantage of large-scale pre-training. We reveal that \textit{simple-finetuning} of existing methods\cite{dust3r_cvpr24,mast3r_arxiv24,wang2025vggt} \textit{does-not} work, whilst our SkyNet \textit{does not face} this issue. 

Our contributions are summarized below:
\begin{enumerate}
    \item \textit{Sky2Ground Dataset}: A large-scale dataset integrating satellite, aerial, and ground-level imagery with \textit{both} synthetic and real components, \textit{providing  annotations} for camera poses and dense depth maps.
    \item \textit{First} benchmark studying existing methods on cross-view localization/reconstruction involving \textit{all three} ground/aerial/satellite imagery.
    \item \textit{SkyNet Architecture}: which outperforms Dust3r/Mast3r/Map-Anything/VGGT on cross-view localization, and requires \textit{only a} single-forward pass.
\end{enumerate}

\section{Related Works}
\label{sec:rw}

\indent \textbf{Cross-View Camera Localization:} requires inferring camera poses (intrinsics/extrinsics) across \textit{aerial, ground and satellite views}. Classical SfM pipelines like Colmap\cite{schoenberger2016mvs,schoenberger2016sfm} face issues in matching-features across such diverse-viewpoints. Additionally, the large-scale (in meters) of captured images, as well as their varying-altitude further exacerbates this problem. Other works \cite{dust3r_cvpr24,mast3r_arxiv24,hu2022beyond} instead learn canonical-transformation between a \textit{pairs of views} via neural-nets. However, they require an additional \textit{global-alignment} phase to map all prediction-pairs onto a \textit{consistent} point-map representation. Recently, VGGT\cite{wang2025vggt} has shown that \textit{a single-forward pass} through transformer is enough to infer camera parameters. \\
\indent In this work, we observe that simply fine-tuning VGGT on combination of \textit{ground/aerial/sat} views \textit{hurts} performance. We propose an alternate method named \textbf{SkyNet}, which we hope is a \textit{small-step} towards fixing this non-trivial problem. \\
\indent \textbf{Cross-View and Novel-View Synthesis:} Recent advances in Neural Radiance Fields (NeRF)~\cite{mildenhall2021nerf} and Gaussian Splatting~\cite{kerbl20233d} have significantly improved cross-view and novel-view synthesis across diverse capture settings. Several methods leverage aerial imagery~\cite{tang2025dronesplat, xiangli2022bungeenerf, zhang2025birdnerf, chen20253d, wu20243d, yuan2025robust, gao2025citygs, liu2024citygaussian, liu2024citygaussianv2} to model large-scale scenes~\cite{somanath2024towards,kikuchi2022future}, while hybrid approaches combine aerial, and ground viewpoints to enhance coverage and geometric stability~\cite{jiang2025horizon, zhang2025crossview, joshi2025unconstrained, ham2024dragon,zhai2017predicting,modi2023occlusions}. Some works~\cite{xiangli2022bungeenerf, liu2024citygaussian} incorporate level-of-detail representations to handle varying scales and scene complexity, whereas others~\cite{zhang2025crossview} fuse separate sub-models for aerial and ground views to improve cross-view consistency. Progressive strategies such as Dragon~\cite{ham2024dragon} gradually integrate images captured from different altitudes to stabilize reconstruction.\\ \indent In contrast, our method performs a unified, full-scale reconstruction that jointly integrates aerial, ground, and satellite observations, demonstrating that improved pose estimation directly translates to more accurate and coherent scene reconstruction.

\begin{table}[ht!]
\vspace{-1em}
\centering
\caption{Comparison with existing datasets. A: Aerial, G: Ground, Sat: Satellite, Syn: Synthetic.}
\resizebox{\columnwidth}{!}{
\begin{tabular}{lcccccccc}
\toprule
\textbf{Dataset} & \textbf{Sites} & \textbf{Imgs}  & \textbf{Syn A} & \textbf{Syn G} & \textbf{Real A} & \textbf{Real G} & \textbf{Sat}  \\
\midrule
KITTI-360~\cite{liao2022kitti}        & 1  & 300k   & \ding{55} & \ding{55} & \ding{55} & \checkmark & \ding{55}  \\
nuScenes~\cite{caesar2020nuscenes}    & 2  & 1.4M  & \ding{55} & \ding{55} & \ding{55}  & \checkmark & \ding{55}  \\
GauU-Scene~\cite{xiong2024gauu}       & 6  & 4.7k   & \ding{55}  & \ding{55} & \checkmark  & \ding{55} & \ding{55}  \\
UC-GS~\cite{zhang2024drone}           & 2  & 1.4k     & \checkmark & \checkmark & \ding{55}  & \ding{55} & \ding{55}  \\
BungeeNeRF~\cite{xiangli2022bungeenerf} & 12 & 3.1k   & \checkmark & \ding{55} & \ding{55}  & \ding{55} & \checkmark  \\
MatrixCity~\cite{li2023matrixcity}    & 1  & 519k      & \checkmark & \checkmark & \ding{55}  & \ding{55} & \ding{55}  \\
AerialMegaDepth~\cite{vuong2025aerialmegadepth}    & 137  & 132k       & \checkmark & \checkmark & \ding{55}  & \checkmark  &  \ding{55} \\
\midrule
\textbf{Ours}                         & \textbf{51} & \textbf{80k}  & \checkmark & \checkmark & \checkmark  & \checkmark & \checkmark \\
\bottomrule
\end{tabular}}
\vspace{-1em}
\label{tab:custom_comparison}
\end{table}

\noindent \textbf{Public Datasets:}  In Tab\ref{tab:custom_comparison}, we compare across existing datasets in the literature. However, we notice them \textit{lacking} in several-aspects. For eg, nuScenes~\citep{caesar2020nuscenes}/ KITTI-360~\citep{liao2022kitti} are autonomous-driving datasets, and lack aerial/satellite views. Similarly, aerial-megadepth\cite{vuong2025aerialmegadepth} contains ground/aerial views, but \textit{lacks} satellite-views. Similarly, synthetically-rendered datasets (using game-engines/unity3d) like MatrixCity, Bungee-Nerf \textit{remain limited} to \textit{synthetic} ground/aerial scenarios. In contrast, our Sky2Ground dataset covers all three ground, aerial, and sat viewpoints, as well as manually-scraped \textit{real} ground/aerial views, indicating its versatility. \\
 \indent To the best of our knowledge there is \textit{no prior-work} that offers: 1) a comprehensive dataset across \textit{all three} viewpoints with \textit{both} real/synthetic images 2) an in-depth study analyzing impact of different viewpoints on camera-localization/rendering.

\section{Sky2Ground Dataset}
We present Sky2Ground: a dataset which consists of images captured from  ground, aerial, and street viewpoints. In Fig \ref{fig:data}, we show few samples from our dataset. It consists of several geographic locations, for eg, \textit{(westminster pier, pyramid du louvre, eros)}. For each location, we have \textit{both} real/synthetic aerial/ground images, as well as satellite imagery captured at varying-altitude levels. Additionally, we provide densely-annotated depth-maps, and camera intrinsics/extrinsics for \textit{each} captured-image. Next, we delve deeper into dataset-statistics.

\label{sec:method}

\begin{table}[H]
\centering
\caption{Sky2Ground dataset statistics.
 A: Aerial, G: Ground, Sat: Satellite, Syn: Synthetic. -: altitude for real is unknown, due to lack of ground-truth poses.}
\resizebox{\columnwidth}{!}{
\begin{tabular}{lccccc}
\toprule
\textbf{Attribute} & \textbf{Sat} & \textbf{Syn A} & \textbf{Real A} & \textbf{Syn G} & \textbf{Real G} \\
\midrule
Modality         & Ortho-Rectified & RGB  & RGB  & RGB & RGB \\
\#Images per site        & 120   & 1080 & 120 & 50--250 & 120\\
Altitude (AGL)   & 1--2 km & 200--800 m & - & 5--80 m & -\\
\bottomrule
\end{tabular}}
\vspace{-1em}
\label{tab:tier_overview}
\end{table}

\begin{figure*}[t!]
  \centering
\vspace{-2.5em}
  \includegraphics[width=0.92\linewidth]{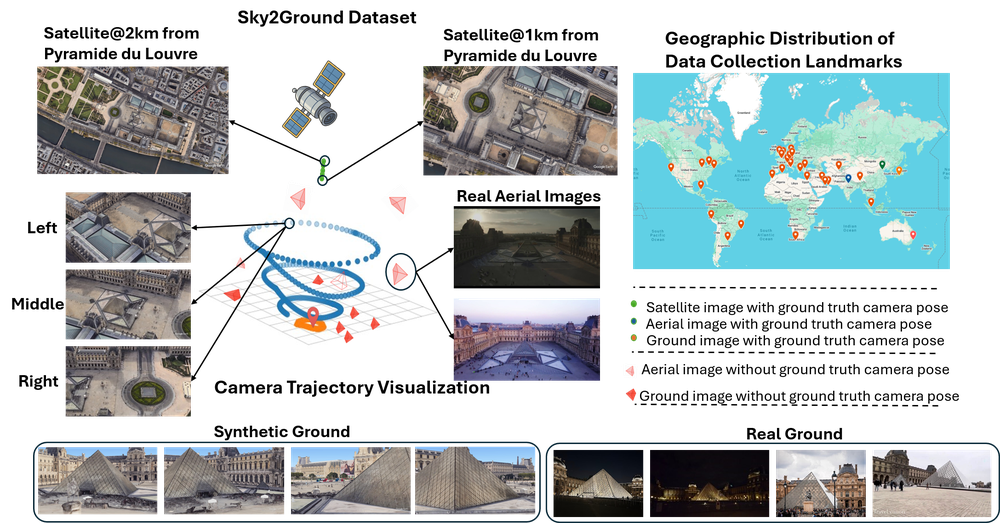}
  \caption{\noindent \textbf{Overview of the Sky2Ground dataset.} 
The middle trajectory illustrates camera poses from one of our collected sites.
Dots indicate ground-truth camera positions for synthetic images, while red frustums represent the estimated camera poses for real images.
The surrounding images showcase example satellite, aerial, and ground views—where the real images demonstrate more diverse illumination conditions and realistic noise.
The top-right map depicts the geographic distribution of all collected landmarks, highlighting the dataset’s global coverage.
}
  \vspace{-2em}
  \label{fig:drone}
\end{figure*}

\indent \textbf{Dataset Statistics:} Sky2Ground consists of $51$ geographical locations, Fig \ref{fig:drone}(right), shows how they span the entire-earth, thereby ensuring diversity. In total, our dataset contains $80k$ images. Tab \ref{tab:tier_overview} provides \textit{individual} stats per-site. Sky2Ground consists of $120$ \textit{synthetic} satellite images, 1080 aerial-images, and $50-250$ ground images. Further, each site contains $120$ real ground/aerial images. In Fig \ref{fig:drone}, we show how the real-images capture weather-conditions, and diverse-lightning (day or night), and altitude levels. Next, we describe the collection-procedure involved in curating our dataset.\\
\indent \textbf{Collection-procedure:}
Sky2Ground consists of both real/synthetic images. We manually-curate \textit{real-imagery} from publicly available Google Maps reviews and YouTube travel vlogs. Next, we filter-them to ensure accurate-coverage of a landmark. Similarly, we collect all the \textit{synthetic data} via Google Earth Studio. However, note that owing to the large degree-of-freedom \textit{(dof)} in 3D, we could capture a particular landmark from \textit{infinite} viewpoints. Therefore, we rely on sampling cameras along different \textit{trajectory-paths} to ensure accurate coverage. Next, we discuss how individual ground, aerial, sat images are curated.\\
\indent For \textit{ground views}, we simulate a single RGB camera performing a circular walk-around of the landmark. For \textit{satellite Views}, we sample images from an altitude of 1-2Km, thereby getting a top-down coverage of the entire scene.  Due to high altitude, the captured images contain perspective-distorations, which we fix via an additional \textit{ortho-rectification} procedure. \\
\indent Finally, for \textit{aerial viewpoints}, we design a virtual triple-camera rig with left, center, and right views at yaw angles of $-20^\circ$, $0^\circ$, and $+20^\circ$, respectively, following a downward helical trajectory. The trajectory begins around $800$ m AGL and descends in stages of $100$–$150$ m, forming three altitude bands: high (800 m), medium (500 m), and low (250 m). Each camera captures $60$/$120$/$180$ frames per band, totaling $360$ images per camera and $1{,}080$ aerial views \textit{per site.}\\
\indent \textbf{Annotation procedure:} By default, google-earth gives us poses of cameras. However, it lacks dense-predictions such as depth-maps/surface-normals. These in turn are required to supervise a 3D-network like VGGT\cite{wang2025vggt}. Therefore, we run dense-MVS pipeline via ColMAP. Each site takes roughly a day-and-half to run with PatchMatch Stereo, and Stereo-Fusion consuming most of our time.

\section{Sky2Ground Benchmark}
Here we present Sky2Ground Benchmark where we evaluate existing models for camera localization/rendering-based tasks. We describe the experimental setup, and then present additional analysis. 

 \begin{figure}[ht!]
    \centering
    \includegraphics[width=0.9\linewidth]{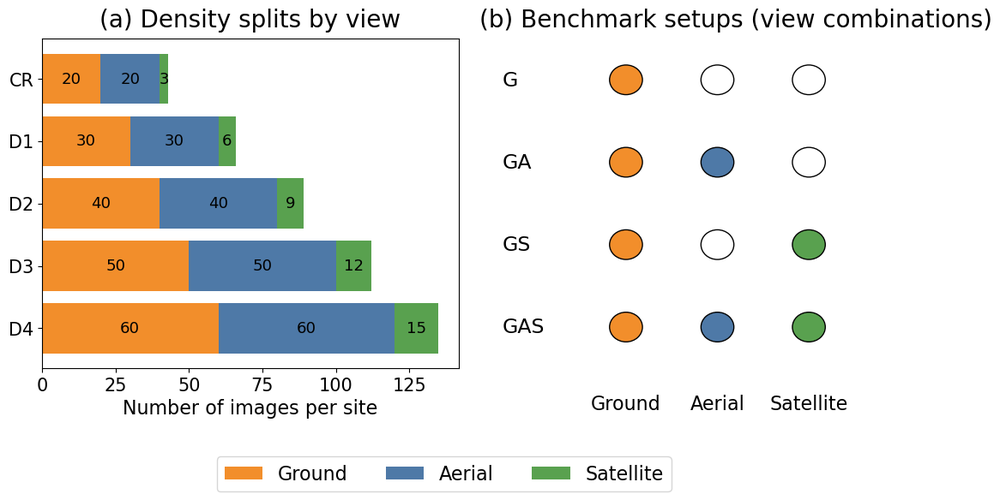}
    \vspace{-0.8em}
    \caption{\textbf{Benchmark splits and modality setups.}
  (a) Image counts per split for synthetic CR - Core, D1 - Dense 1, D2 - Dense 2, D3 - Dense3 and D4 - Dense 4, across ground, aerial, and satellite views.
  (b) View combinations used in each benchmark setup: Ground (G), Ground+Aerial (GA), Ground+Satellite (GS), and Ground+Aerial+Satellite (GAS).
  }
     \vspace{-2.0em}
    \label{fig:benchmark split}
\end{figure}

\indent \textbf{Experimental setup:}
In Fig.~\ref{fig:benchmark split}, we show that our dataset comprises of five synthetic splits (CR, D1–D4). Here D1–D4 provide progressively denser-image coverage. Note that core images, (CR) are \textit{common} to all the splits, i.e. (CR, D1-D4). We analyze existing methods on $D2$ setup. We \textit{evaluate} on a subset of $D2$ images, i.e. the core-setup $CR$. This ensures that although the \textit{total number} of images across $D1-D4$ changes, evaluation setup \textit{stays consistent}. Additionally, we evaluate various gaussian-splatting methods for scene-rendering. Therefore, our analysis covers \textit{both} localization/rendering.\\
\indent \textbf{Models.} We benchmark a wide variety of \textit{learning-based} methods for camera-localization. For eg, we evaluate \textit{pair-wise view processing} nets like Dust3r/Mast3r/MapAnything. Additionally, we evaluate \textit{VGGT} which infers camera-poses in a single feed-forward. Additionally, we benchmark classical SfM techniques like COLMAP, and provide more results in the supplementary. For rendering, we evaluate models like %
2DGS, 3DGS, and 3DGS-MCMC, which span 2D splatting, 3D Gaussian scene representations, and an MCMC-refined 3D variant.\\
\indent \textbf{Metrics:} We evaluate camera-localization performance via Relative Rotation Accuracy (RRA) and Relative Translation Accuracy (RTA)~\citep{Jin2020rrarta}. More specifically, we threshold at $5^\circ$, and report RRA@5$^\circ$/ RTA@5$^\circ$. This denotes, how well the model localizes cameras within $5^\circ$ of each-other. We quantitatively evaluate photometric/perceptual quality of 3D-renderings via PSNR and DreamSim metrics. A higher value of PSNR is assumed better, whereas for DreamSim \textit{lower-score} is better.

\subsection{Zero-shot Analysis}
Here, we first discuss about \textit{zero-shot} localization performance of existing models, and variation as number of input views changes. Next, we discuss rendering results. 

\begin{figure}[ht!]
    \centering
    \vspace{-1em}
\includegraphics[width=0.95\columnwidth]{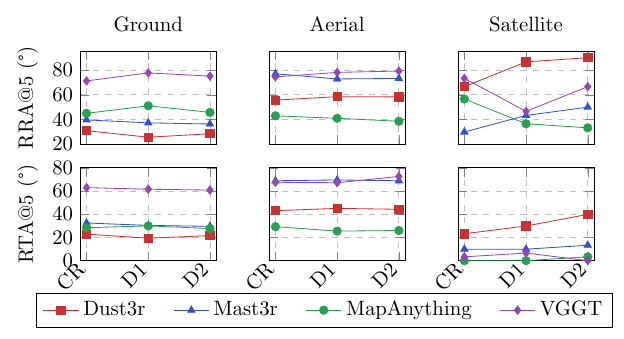}
    \caption{
    Comparison of \textbf{RRA@5} and \textbf{RTA@5} metrics for four methods 
    (Dust3r, Mast3r, Map Anything, and VGGT).
    }
    \label{fig:rra_rta_lineplots}
    \vspace{-1em}
\end{figure}

\indent \textbf{Models `suffer' when number of views are less}: In Fig.~\ref{fig:rra_rta_lineplots}, we plot how $RRA/RTA$ varies  as the number of input views increases $core\rightarrow dense1 \rightarrow dense2$. We observe models perform \textit{worse} on sparse scenarios, i.e. \textit{core}. \\
\indent \textbf{VGGT performs best on Ground/Aerial setup:} We note that VGGT (purple-curve) performs best on Ground/Aerial scenarions. One notable thing is that performance of VGGT \textit{does not fluctuate} much as compared to Dust3r/Mast3r/Map-Anything, indicating that it does well in \textit{sparse-scenarios.}\\
\indent \textbf{Pair-wise view networks like Dust3r do well on satellite}: In Fig.~\ref{fig:rra_rta_lineplots}, we note that Dust3r achieves highest performance on localizing \textit{satellite} images (red curve). We believe that this is due to \textit{global-alignment} enforced in dust3r. Satellite-views contain a lot of mutually-common points, and an explicit matching step ensures proper alignment. However, we note that Dust3r/Mast3r \textit{don't} perform well on \textit{ground} scenarios. \\

\begin{figure}[ht!]
    \centering
    \vspace{-2em}
\includegraphics[width=\columnwidth]{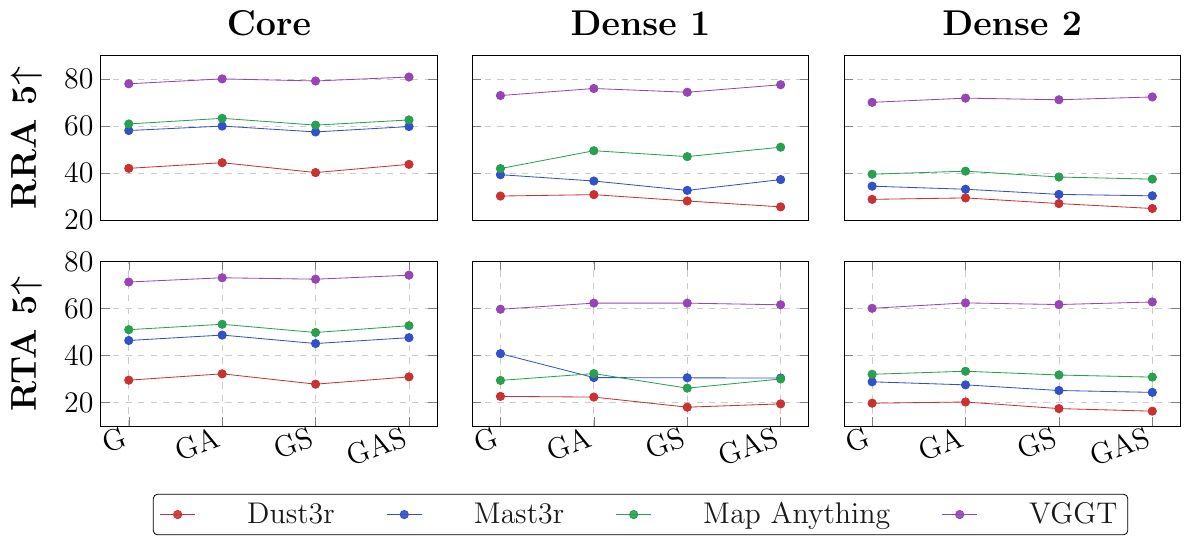}
    \caption{
    Comparison of models across view combinations.
    }
    \vspace{-1em}
    \label{fig:g_ga_gs_gas}
\end{figure}

\noindent \textbf{Satellite images pose a `significant-problem':} One expectation might be that models  \textit{ should perform better} on ground+aerial+satellite (GAS) setup than ground+aerial setup, because adding satellite provides more information to the neural-net. In Fig\ref{fig:g_ga_gs_gas}, we surprisingly find that this \textit{is not} the case. Instead, we observe \textit{drops} in performance. Therefore, processing \textit{satellite} modality remains a challenge.\\

\begin{figure}[ht!]
\vspace{-2em}
  \centering
   \includegraphics[trim={0, 0, 0, 0cm}, clip, width=0.7\linewidth]{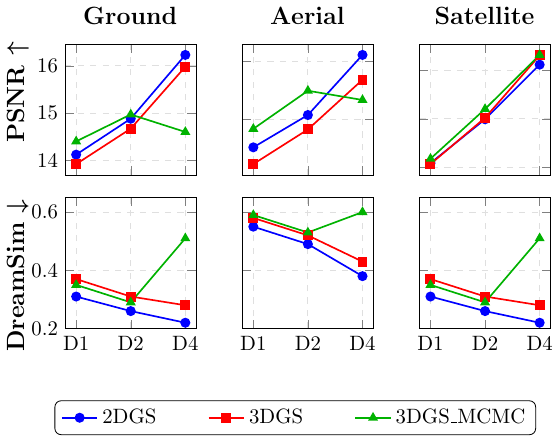}
   \caption{\textbf{Comparison of reconstruction quality across ground/aerial/satellite.}: We report PSNR$\uparrow$ and DreamSim$\downarrow$ (lower is better). All methods benefit from increased camera density. 2DGS consistently achieves the best perceptual quality. 
  }
   \label{fig:dreamsim}
   \vspace{-1em}
 \end{figure}

\noindent \textbf{2D-GS consistently gives best rendering results across ground/aerial/satellite:} Earlier, we noticed that VGGT obtained the \textit{best} performance out of all localization methods. We use VGGT zero-shot predicted poses, to initialize various gaussian splatting methods (2DGS, 3DGS and 3DGS-MCMC). We evaluate performance as number of input views increases from $D_1 \rightarrow D_2\rightarrow D_4$. The scaling-law (slope) suggests that 2D-Gaussian splatting consistently performs better. 

\textbf{Adding Real images harms rendering and yields noisy reconstructions:} 
Tab.~\ref{tab:psnr_dreamsim} shows that introducing real images into the training mix consistently degrades rendering quality across GS types and viewpoints.
Real-world lighting variation, sensor noise, clutter, and textures absent from synthetic data create a strong domain mismatch, making it difficult for the model to blend real and synthetic imagery.

\begin{figure*}[t!]
    \vspace{-2em}
    \centering
    \includegraphics[width=\textwidth]{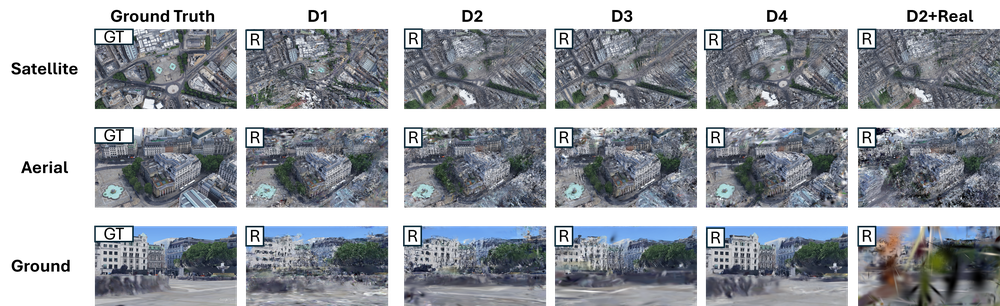}
    \caption{
        Rendering results across satellite, aerial, and ground viewpoints.
        Each row shows a different view, with the leftmost column showing the ground-truth (GT) image followed by model renderings (R).
        From D1 to D4, the input density increases, and real images are incorporated with D2, introducing additional noise and domain mismatch.
        In particular, the ground-view renderings in D2 + Real contain real-image features that do not exist in the synthetic input distribution, highlighting the challenge of mixing synthetic and real sources.
    }
    \label{fig:synthesis_results}
\end{figure*}

Fig.~\ref{fig:synthesis_results} shows that increasing density from D1 to D4 produces sharper contours and more complete details, while mixing in 30 aerial and 30 ground real images at D2 introduces noise and domain mismatch, especially in ground views where real features absent from synthetic data begin to appear.

\begin{table}[t]
\centering
\caption{Rendering results across synthetic and real-image setups. We report PSNR and DreamSim for each GS type (2DGS, 3DGS, 3DGS\textsubscript{\tiny MCMC}) across aerial, street, and satellite viewpoints.}
\label{tab:psnr_dreamsim}
\setlength{\tabcolsep}{1.5pt}
\small
\resizebox{0.48\textwidth}{!}{
\begin{tabular}{l l ccc ccc}
\toprule
& & \multicolumn{3}{c}{PSNR $\uparrow$} & \multicolumn{3}{c}{DreamSim $\downarrow$} \\
\cmidrule(lr){3-5}\cmidrule(lr){6-8}
Data & GS Type & Aerial & Ground & Satellite & Aerial & Ground & Satellite \\
\midrule
\multirow{3}{*}{Synthetic}
    & 2D                    & 12.1 & 14.9 & 13.0 & 0.49 & 0.26 & 0.19 \\
    & 3D                    & 11.8 & 14.7 & 13.0 & 0.52 & 0.31 & 0.17 \\
    & 3D\textsubscript{\tiny MCMC}
                              & 12.5 & 15.0 & 13.3 & 0.53 & 0.29 & 0.21 \\
\midrule
\multirow{3}{*}{Syn+Real}
    & 2D                    & 11.1 & 13.9 & 12.6 & 0.61 & 0.40 & 0.26 \\
    & 3D                    & 11.1 & 13.4 & 12.6 & 0.64 & 0.46 & 0.28 \\
    & 3D\textsubscript{\tiny MCMC}
                              & 11.2 & 12.8 & 13.4 & 0.79 & 0.71 & 0.68 \\
\bottomrule
\vspace{-3em}
\end{tabular}
}
\end{table}

\begin{table*}[ht!]
\centering
\scriptsize
\setlength{\tabcolsep}{4pt}
\renewcommand{\arraystretch}{1.05}
\caption{\textbf{Localization results across Ground+Aerial+Satellite setup.}
ZS = Zero-Shot, A-MD = Aerial-MegaDepth. RRA@5 / RTA@5 measure rotation / translation accuracy within 5° / 5 m. RRA Avg and RTA Avg denote mean performance across all domains.}
\resizebox{0.9\textwidth}{!}{
\begin{tabular}{lcccccccccc}
\toprule
\textbf{Method} & \textbf{Data} &
\multicolumn{2}{c}{\textbf{Ground}} &
\multicolumn{2}{c}{\textbf{Satellite}} &
\multicolumn{2}{c}{\textbf{Aerial}} &
\multicolumn{2}{c}{\textbf{Avg}} \\
\cmidrule(lr){3-4} \cmidrule(lr){5-6} \cmidrule(lr){7-8} \cmidrule(lr){9-10}
 &  & RRA@5 & RTA@5 & RRA@5 & RTA@5 & RRA@5 & RTA@5 & RRA & RTA \\
\midrule
Dust3r & ZS & 28.5\textsubscript{\tiny$\pm$0.71} & 21.6\textsubscript{\tiny$\pm$0.53} & 90.0\textsubscript{\tiny$\pm$1.92} & 39.9\textsubscript{\tiny$\pm$0.84} & 58.3\textsubscript{\tiny$\pm$1.37} & 44.3\textsubscript{\tiny$\pm$1.18} & 58.9\textsubscript{\tiny$\pm$1.23} & 35.3\textsubscript{\tiny$\pm$0.79} \\
Mast3r & ZS & 36.4\textsubscript{\tiny$\pm$0.86} & 29.6\textsubscript{\tiny$\pm$0.77} & 50.0\textsubscript{\tiny$\pm$1.04} & 13.3\textsubscript{\tiny$\pm$0.28} & 73.2\textsubscript{\tiny$\pm$1.49} & 68.9\textsubscript{\tiny$\pm$1.26} & 53.2\textsubscript{\tiny$\pm$1.07} & 37.3\textsubscript{\tiny$\pm$0.91} \\
VGGT & ZS & 75.1\textsubscript{\tiny$\pm$1.83} & 60.9\textsubscript{\tiny$\pm$1.29} & 66.6\textsubscript{\tiny$\pm$1.42} & 0.0\textsubscript{\tiny$\pm$0.00} & 79.2\textsubscript{\tiny$\pm$1.67} & 72.6\textsubscript{\tiny$\pm$1.23} & 73.6\textsubscript{\tiny$\pm$1.38} & 44.5\textsubscript{\tiny$\pm$0.95} \\
\hline
Dust3r & A-MD & 25.5\textsubscript{\tiny$\pm$0.64} & 23.2\textsubscript{\tiny$\pm$0.71} & 83.3\textsubscript{\tiny$\pm$1.78} & 29.9\textsubscript{\tiny$\pm$0.82} & 83.6\textsubscript{\tiny$\pm$1.74} & 64.8\textsubscript{\tiny$\pm$1.35} & 64.1\textsubscript{\tiny$\pm$1.27} & 39.3\textsubscript{\tiny$\pm$0.92} \\
Mast3r & A-MD & 29.1\textsubscript{\tiny$\pm$0.73} & 24.1\textsubscript{\tiny$\pm$0.61} & 96.6\textsubscript{\tiny$\pm$2.31} & 26.6\textsubscript{\tiny$\pm$0.68} & 77.0\textsubscript{\tiny$\pm$1.59} & 94.5\textsubscript{\tiny$\pm$2.18} & 67.6\textsubscript{\tiny$\pm$1.33} & 48.4\textsubscript{\tiny$\pm$0.97} \\
\hline
Dust3r & Sky2Ground & 32.4\textsubscript{\tiny$\pm$0.68} & 27.1\textsubscript{\tiny$\pm$0.63} & 79.4\textsubscript{\tiny$\pm$1.54} & 36.3\textsubscript{\tiny$\pm$0.87} & 83.8\textsubscript{\tiny$\pm$1.79} & 67.0\textsubscript{\tiny$\pm$1.16} & 65.2\textsubscript{\tiny$\pm$1.05} & 43.5\textsubscript{\tiny$\pm$0.94} \\
Mast3r & Sky2Ground & 33.2\textsubscript{\tiny$\pm$0.66} & 27.9\textsubscript{\tiny$\pm$0.59} & 81.0\textsubscript{\tiny$\pm$1.47} & 37.2\textsubscript{\tiny$\pm$0.81} & 85.1\textsubscript{\tiny$\pm$1.71} & 68.3\textsubscript{\tiny$\pm$1.12} & 69.4\textsubscript{\tiny$\pm$1.08} & 47.5\textsubscript{\tiny$\pm$0.89} \\
VGGT & Sky2Ground & 50.0\textsubscript{\tiny$\pm$1.25} & 46.1\textsubscript{\tiny$\pm$1.17} & 86.6\textsubscript{\tiny$\pm$1.94} & 53.3\textsubscript{\tiny$\pm$1.41} & 29.7\textsubscript{\tiny$\pm$0.61} & 31.5\textsubscript{\tiny$\pm$0.69} & 55.4\textsubscript{\tiny$\pm$1.09} & 43.6\textsubscript{\tiny$\pm$0.97} \\
\rowcolor{gray!20}
SkyNet & Sky2Ground & 76.7\textsubscript{\tiny$\pm$1.52} & 64.2\textsubscript{\tiny$\pm$1.27} & 88.9\textsubscript{\tiny$\pm$1.93} & 57.3\textsubscript{\tiny$\pm$1.14} & 84.0\textsubscript{\tiny$\pm$1.48} & 78.1\textsubscript{\tiny$\pm$1.06} & \textbf{83.2}\textsubscript{\tiny$\pm$1.37} & \textbf{66.5}\textsubscript{\tiny$\pm$0.98} \\
\bottomrule
\end{tabular}
\vspace{-7em}
}
\label{tab:gas_results}
\end{table*}

\subsection{Fine-Tuning-Analysis using Sky2Ground}
\label{sec:fine_tuning}
Earlier we noted that adding satellite during \textit{zero-shot inference} harms performance. In this section, we aim to study if \textit{simply adding} satellite during training, i.e. jointly fine-tuning existing models could work. To facilitate a systematic understanding, we first train on Ground-aerial (GA) data, and then on Ground/Aerial/Satellite data. Out of $51$ sites in Sky2Ground, we use $41$ for training, and remaining $10$ for evaluation.

\indent \textbf{Fine-Tuning with Ground-Aerial Data improves performance:} Tab \ref{tab:gas_results}, shows that training Dust3r on ground-aerial data improves on avg from $58.9\rightarrow 64.1 (RRA), 35.3 \rightarrow 39.3 (RTA)$. Similarly, Mast3r improves from $ 53.2 \rightarrow 67.6 (RRA), 37.3 \rightarrow 48.4 (RTA)$. Next, we study the effect of adding `satellite'.\\
\indent \textbf{Adding satellite improves `pairwise-view' nets like Dust3r/Mast3r} We note that adding satellite improves performance of nets like Dust3r/Mast3r from $64.1 \rightarrow 65.2 \text{ (RRA avg})$, $67.6 \rightarrow 69.4           \text{ (RTA avg)}$ respectively. A similar trend is observed for RTA. An inductive-bias common to \textit{both} Dust3r/Mast3r is that both process a \textit{pair of views} together, which \textit{might} explain this behaviour. However, choosing a pair out of $N$ total views requires $\binom{N}{2} \approx O(N^2)$ feed-forwards through the net, thereby making the entire pipeline \textit{extremely} slow. Next, we study the effect of adding `satellite' during training to \textit{single} feed-forward nets. \\
\indent \textbf{Adding satellite to `single-feedforward' nets like VGGT \textit{severely} hurts performance.}: In Tab\ref{tab:gas_results}, we observe that VGGT suffers a \textit{severe} drop, i.e. $73.6 \rightarrow 55.4$ (\textcolor{red}{-18.2\% RRA}), $44.5 \rightarrow 43.6$ (\textcolor{red}{-0.9\% RTA}). The above analysis leads to a \textit{counter-intuitive} insight. Adding satellite \textit{hurts} VGGT but not Dust3r/Mast3r. If extreme-distribution shift across ground/aerial/sat views \textit{were} the \textit{only} cause, we should have observed consistent-drops across \textit{all} methods. This leads us to question: \uline{`How can we further-improve  single-feedforward nets like VGGT?'}. Building upon VGGT makes sense because we can then \textit{bypass} classical SfM pipelines (aka COLMAP), global-alignment stage, and perform E2E optimization.

\vspace{-1em}
\section{Improving localization with satellite imagery}
As discussed in previous section \ref{sec:fine_tuning}, addition of satellite does not work even with fine-tuning. Inspired by this observation, here we present \textbf{SkyNet}, an architecture for cross-view camera localization. First, we formalize our problem statement and describe preliminaries.

\begin{figure*}
  \centering
  \vspace{-2em}
  \includegraphics[trim={0, 0, 0, 0cm}, clip, width=1\linewidth]{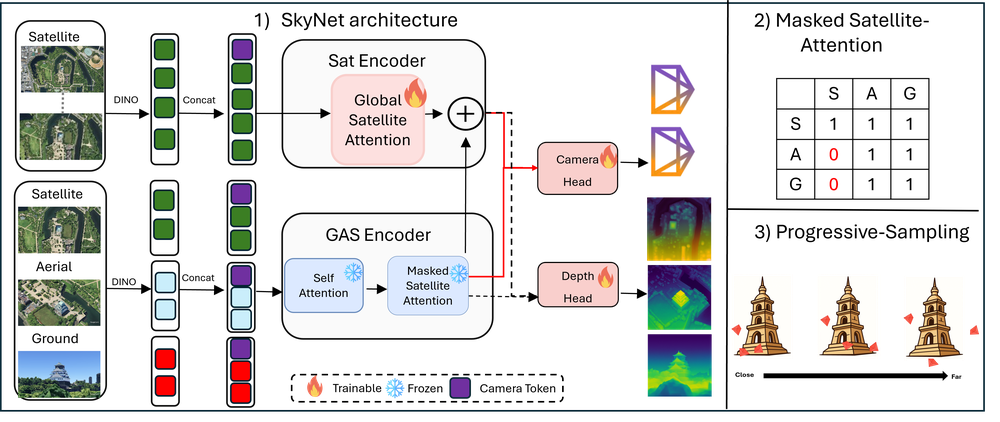}
  \caption{\textbf{SkyNet:} An architecture for cross-view camera localization. SkyNet consists of two encoders, 1) a Sat-Encoder processes input satellite images, and a GAS-Encoder processes ground/aerial/satellite views. Our model first patchifies the input images into tokens by DINO, and appends camera tokens for camera
prediction. GAS-encoder then alternates between self-attention and Masked-Satellite Attention. Camera/Depth heads collect satellite tokens from Sat-encoder, ground-aerial tokens from GAS-encoder and predict camera-poses and depth-maps respectively. 2) Masked-Satellite Attention \textit{restricts} aerial/ground from interacting with satellite modality. 3) Progressive-sampling gradually samples far-away cameras as training proceeds. 
  }
  \label{fig:skynet_v1}
  \vspace{-1.5em}
\end{figure*} 

\indent \textbf{Problem statement:} Given $N$ RGB images $I= \{ I_g, I_a, I_s\}_{N}$ consisting of $\{I_g\}$ ground-views, $\{I_a\}$ aerial-views and $\{I_s\}$ satellite-views, our objective is to predict the corresponding camera parameters $\{ \boldsymbol{g}_i \in \mathbb{R}^9\}^N_{i=1}$ and depth map $\{ D_i \in \mathbb{R}^{H \times W} \}^N_{i=1}$. Following VGGT\cite{wang2025vggt}, we represent $g_i$ in quaternion notation.  \\
\indent  \textbf{Preliminaries on VGGT:} Each of the $N$ input-images is first passed through a dino encoder. The resultant features are then concatenated with additional camera tokens $\mathbf{t}^{\boldsymbol{g}}_i \in \mathbb{R}^{1 \times C'}$ and four register tokens $\mathbf{t}^R_i \in \mathbb{R}^{4 \times C'}$. The camera/register tokens of first input view ($\mathbf{t}^{\boldsymbol{g}}_1 := \bar{\mathbf{t}}^{\boldsymbol{g}}, \ \mathbf{t}^R_1 := \bar{\mathbf{t}}^R$) are set to a \textit{different} set of learnable tokens $(\bar{\mathbf{t}}^{\boldsymbol{g}}, \bar{\mathbf{t}}^R)$ than those of all other frames ($\mathbf{t}^{\boldsymbol{g}}_i := \bar{\bar{\mathbf{t}}}^{\boldsymbol{g}}, \ \mathbf{t}^R_i := \bar{\bar{\mathbf{t}}}^R, \ i \in [2, \dots, N]$), which are also learnable.\\
\indent VGGT consists of $L$ layers of alternating frame-attention and global-attention. First, frame-attention allows tokens of each frame to interact among themselves. Next, in global-attention, information across different-views is aggregated, and aligned in the latent space. As in Sec\ref{sec:fine_tuning}, we notice two-issues with fine-tuning VGGT 1) It \textit{disturbs} the pre-trained VGGT weights, thereby losing the advantage of large-scale pre-training 2) Global Frame-attention \textit{forces} ground/aerial views to exchange information with satellite, but distribution-shift \textit{hurts} performance. 

\subsection{SkyNet}
Here we propose SkyNet, an architecture that processes input data in two parallel-streams. SkyNet consists of two components 1) a ground-aerial-sat (GAS)-encoder 2) a sat-encoder. \\
\indent \textbf{GAS-Encoder:} The Ground-Aerial-Satellite (GAS-encoder) consists of $L$ blocks. In each block, \textit{all} input views first undergo a self-attention. Next, we perform \textit{restricted-attention} termed as \uline{M}asked-\uline{S}atellite-\uline{A}ttention \textbf{(MSA)}. In Fig \ref{fig:skynet_v1}, MSA \textit{restricts}  \textit{ground/aerial} views  from interacting with \textit{sat}. However, \textit{sat} still interact with aerial/gnd views. The resulting attention values
$V = [v_G, v_A, v_S]$ are passed to $L-1$ identical blocks.  The self-attention, and MSA block are initialized with pre-trained VGGT weights, and kept-frozen. After each block, sat features $v_S$ are fed to the sat-encoder. \\
\indent \textbf{Sat-Encoder:} Given multiple satellite images $I_s$, we feed-forward them through a dino encoder. The output tokens first undergo a global-attention, where all the sat-views interact with each other, resulting in the latent $z$. Next, we update $z = z +v_s$, where $v_s$ are the refined \textit{sat} features coming from the GAS-encoder. \\
\indent \textbf{Camera/Depth-Map Predictions:} 
The camera parameters $(\hat{\boldsymbol{g}}^i)_{i=1}^N$ are predicted from the output camera tokens $({\hat{\mathbf{t}}}^\bg_i)_{i=1}^N$ using four additional self-attention layers followed by a linear layer. 
This camera-head decodes satellite-poses from the satellite-encoder, and aerial/ground poses from the GAS encoder. Additionally, we leverage a DPT head to output dense features $T_i \in \mathbb{R}^{C \times H \times W}$. Both camera/depth-head are \textit{shared} across GAS encoder/satellite encoder. \\
 \indent \textbf{Losses:} We set the number of transformer layers as $L=24$. We train SkyNet end-to-end via a multi-task loss 
$\mathcal{L} = \mathcal{L}_{\text{cam, sat}} + \alpha \mathcal{L}_{\text{cam, gnd/aerial}} + \mathcal{L}_{\text{depth}}$, where we empirically set $\alpha = 0.4$. 
The camera loss, depth loss, and other hyperparameters assume similar implementation/initialization as VGGT\cite{wang2025vggt}.

\vspace{-1em}
\begin{table*}[t!]
\centering
\scriptsize
\setlength{\tabcolsep}{2pt}
\renewcommand{\arraystretch}{1.05}
\begin{minipage}{0.48\linewidth}
\centering
\vspace{-15em}
\caption{\textbf{Localization results across Ground+Satellite setup.}
RRA@5 / RTA@5 denote rotation and translation accuracy within 5° / 5 m.
ZS = Zero-Shot.}
\vspace{-4em}
\begin{tabular}{lcccccccc}
\toprule
\textbf{Method} & \textbf{Data} &
\multicolumn{3}{c}{\textbf{Ground}} &
\multicolumn{3}{c}{\textbf{Satellite}} &
\textbf{Overall Avg} \\
\cmidrule(lr){3-5} \cmidrule(lr){6-8}
 &  & RRA@5 & RTA@5 & Avg & RRA@5 & RTA@5 & Avg &  \\
\midrule
Dust3r & ZS & 29.2 & 19.1 & 24.2 & 90.0 & 56.6 & 73.3 & 48.7 \\
Mast3r & ZS & 32.8 & 31.5 & 32.2 & 73.3 & 33.3 & 53.3 & 42.7 \\
VGGT   & ZS & 70.2 & 61.6 & 65.9 & 76.6 & 3.3  & 40.0 & 52.9 \\
\midrule
VGGT   & Sky2Ground & 34.5 & 40.2 & 37.4 & 83.3 & 33.3 & 58.3 & 47.8 \\
\rowcolor{gray!20} 
\textbf{SkyNet} & \textbf{Sky2Ground} & 71.3 & 64.9 & 68.1 & 88.0 & 36.3 & 62.2 & \textbf{65.1} \\
\bottomrule
\vspace{-24em}
\end{tabular}

\label{tab:gs_results}
\end{minipage}
\hfill
\begin{minipage}{0.48\linewidth}
\centering
\caption{\textbf{Ablation of SkyNet variants.}
Effect of MSA (Masked Satellite Attention), CA-CS (Curriculum-Aware Camera Sampling), and P-VS (Progressive View-Sampling).}
\begin{tabular}{l|ccccc l}
\toprule
\textbf{Method} & \textbf{GAS-Enc.} & \textbf{Sat-Enc.} & \textbf{MSA} & \textbf{CA-CS} & \textbf{P-VS} & \textbf{Avg} \\
\midrule
VGGT (trained) & A+G+S & - & \xmark & \xmark & \xmark & 47.8 \\
\hline
VGGT-ZS        & A+G+S & - & \xmark & \xmark & \xmark & 52.9 \\
\hline
SkyNet (Ours)  & A+G & S & \xmark & \xmark & \xmark & 53.1 \textcolor{mypink}{0.2$\uparrow$}\\
               & A+G+S & S & \xmark & \xmark & \xmark & 53.8 \textcolor{mypink}{0.9$\uparrow$}\\
\cline{2-7}
               & A+G+S & S & \xmark & \checkmark & \xmark & 54.3 \textcolor{mypink}{1.4$\uparrow$}\\
               & A+G+S & S & \xmark & \xmark & \checkmark & 61.1 \textcolor{mypink}{7.3$\uparrow$}\\
               & A+G+S & S & \checkmark & \xmark & \xmark & 62.7 \textcolor{mypink}{8.2$\uparrow$}\\
\cline{2-7}
               & A+G+S & S & \checkmark & \checkmark & \xmark & 63.8 \textcolor{mypink}{10.9$\uparrow$}\\
               & A+G+S & S & \checkmark & \xmark & \checkmark & 64.9 \textcolor{mypink}{12.0$\uparrow$}\\
\cline{2-7}
\rowcolor{gray!20}
               & A+G+S & S & \checkmark & \checkmark & \checkmark & 
               65.1 \textcolor{mypink}{12.2$\uparrow$}\\
\bottomrule
\end{tabular}
\vspace{-2em}
\label{tab:skynet_ablation}
\end{minipage}

\end{table*}

\subsection{Cross-View Training strategies}
\indent As humans, we learn how to solve a particular problem by solving easier instances first, and gradually increasing the level of difficulty. This has been known as `curriculum-learning' in prior-literature. Inspired by this, we propose two training strategies for cross-view localization 1) \uline{C}urriculum \uline{A}ware \uline{C}amera-\uline{S}ampling (\textbf{CA-CS}), 2) \uline{P}rogressive-\uline{V}iew \uline{S}ampling \textbf{(P-VS)}.\\
 \indent \textbf{CA-CS sampling:}  During initial epochs, we sample cameras that lie closer to one other. As training proceeds, we gradually sample cameras far away from one another, for eg, non-overlapping cameras. Consider two-views with extrinsics $(R_1,R_2)$, and $(t_1, t_2)$. We estimate rotation/translation distance $d_R/d_T$ as:
\begin{equation}
d_R = \tfrac{1}{\pi}\arccos\!\left(\tfrac{\operatorname{Tr}(R_1^{\top}R_2)-1}{2}\right), \quad
d_T = \|t_1 - t_2\|_2,
\end{equation}
\noindent where, $D(I_1, I_2) = d_R + \lambda_t d_T$, where we set $\lambda_t = 0.5$, balances the rotation and translation terms. We pre-compute and cache pair-wise distances of all the cameras in a scene. For a given ground-view $I_1$, all the other views are sorted in increasing order of distance, and sampled with equal offset w.r.t each other. Our distance-metric doesn't involve explicit-point/feature-matching as in \cite{dust3r_cvpr24,mast3r_arxiv24}, and runs even on a CPU.\\
\indent \textbf{P-VS sampling:} Intuitively, \textit{ground/sat} views are extreme-viewpoints. However, aerial-images can serve as a `bridge' connecting these  modalities together\cite{vuong2025aerialmegadepth}. During training, we sample $N$ total images, where $N=N_a+N_g+N_s$ images, $N_a,N_g,N_s$ being number of ground/aerial/sat images. In beginning of our training schedule, we sample \textit{more number} of aerial-views $(N_a\approx N)$. Towards the end, we only retain ground/satellite images ($N_a\approx 0$). Therefore, our SkyNet gradually transitions from an easier problem, (localizing ground/aerial/sat views) to harder scenarios (ground/sat only).

\subsection{Results}

We evaluate SkyNet on \textit{dense2} setup, and consider \textit{both} ground/aerial/satellite , as well as ground/satellite setup. Note that ground/satellite is an \textit{extremely-realistic} scenario since aerial-imagery might not be available in certain-regions, for eg, forests, and war-zones.  We report both RRA@5 and RTA@5, with standard-deviation across $10$ sites in Sky2Ground. In Tab\ref{tab:gas_results}, we observe that SkyNet achieves a sota score of $83.2\%$(\textcolor{darkgreen}{+9.6\% RRA}), $83.2\%$(\textcolor{darkgreen}{+22.0\% RTA}). In Tab\ref{tab:gs_results}, we report $avg = (\frac{RRA@5 + RTA@5}{2})$. We observe our SkyNet improves VGGT even further to $65.1$ on average, marking improvement of (\textcolor{darkgreen}{+12.2\% }).

\subsection{Ablations on SkyNet}
In Tab \ref{tab:skynet_ablation}, we discuss how different components in SkyNet perform in isolation w.r.t to each other. \\
\indent \textbf{Processing satellite along with aerial/ground images helps.} Processing \textit{sat} images in \textit{both} Ground-Aerial encoder, and Sat-encoder fares better than processing satellite only in Sat-encoder (+0.9 vs +0.2). \\
\indent \textbf{Masked Sat-Attention is the most contributing component:} In Tab\ref{tab:skynet_ablation}, using Masked-Sat attention results in highest increase of  $8.2$. Masking constraint ensures that satellite-tokens are refined by both ground/aerial tokens. Furthermore, we retain strong zero-shot performance of VGGT in ground/aerial case, since ground-aerial tokens \textit{never} interact with input satellite. \\
\indent \textbf{Progressive-View Sampling (P-VS) is the best training-strategy:} Training SkyNet with \textit{only} P-VS obtains $+7.3$, whereas with CA-CS only improves by $1.4$. This shows that decreasing aerial-views during training is the better strategy.\\
\indent \textbf{Adding satellite features from VGGT encoder$\rightarrow$ Sat-encoder outperforms naive cross-attention.} We perform an additional ablation, where instead of `adding' satellite-tokens from the Ground-Aerial encoder, we replace it with cross-attention\cite{carion2020end}. Avg performance \textit{drops} from $65.1$ to $59.7$, indicating that cross-attention is not the best choice.\\
\indent \textbf{SkyNet's improvements are not merely due to increasing-parameters:} We create a flop-matched baseline by increasing VGGT to additional $24$ layers. This ensures that parameter count becomes \textit{identical} to proposed SkyNet. The performance improves from $47.8$ to $50.4$. However, it still lies below VGGT's zero-shot performance $(52.9)$, as well as SkyNet's results $(65.1)$, indicating that improvements are not \textit{merely} due to increasing parameters, but due to MSA/CA-CS/P-VS components. \\
\vspace{-2em}
\section{Conclusion}
We introduced the problem of jointly localizing cameras across ground, aerial, and satellite views. We presented Sky2Ground, a dataset spanning both real and synthetic imagery, and showed that naïve fine-tuning of existing models is hindered primarily by the distribution shift introduced by satellite views. To address this, we proposed SkyNet, a cross-view extension of VGGT that leverages curriculum-inspired strategies such as progressive view and camera sampling. We sincerely hope that Sky2Ground may serve as a foundation for future research.\\

\noindent \textbf{Limitations:} Our method is inherently two-stage, in first-stage we predict-poses, and then rely on gaussian-splat for rendering. Exploring \textit{unified-models} capable of predicting poses/gaussian-parameters remains an interesting direction. \\

\noindent\textbf{Acknowledgments}
\label{sec:acknowledgments}
This work was supported by Intelligence Advanced Research Projects Activity (IARPA) via Department of Interior/Interior Business Center (DOI/IBC) contract number 140D0423C0074. The U.S. Government is authorized to reproduce and distribute reprints for Governmental purposes, notwithstanding any copyright annotation thereon. Disclaimer: The views and
conclusions contained herein are those of the authors and should not be interpreted as necessarily representing the official policies or endorsements, either expressed or implied, of IARPA, DOI/IBC, or the U.S. Government.

{
    \small
    \bibliographystyle{ieeenat_fullname}
    \bibliography{bibliography}
}

\newpage 

\clearpage

\setcounter{page}{1}
\maketitlesupplementary
  \setcounter{section}{0}
\renewcommand{\thesection}{\Alph{section}}
\renewcommand{\thesubsection}{\thesection\arabic{subsection}}
\begin{figure*}[ht!]
\begin{lstlisting}
# Sky2Ground Satellite Collection Script

# 1. Convert (lat, lon) -> tile coordinates
def latlon_to_tileXY(lat, lon, zoom):
    lat_rad = math.radians(lat)
    n = 2 ** zoom
    x_tile = int((lon + 180.0) / 360.0 * n)
    y_tile = int(
        (1 - math.log(math.tan(lat_rad) + 1 / math.cos(lat_rad)) / math.pi) / 2 * n
    )
    return x_tile, y_tile

# 2. TileXY -> Quadkey
def tileXY_to_quadkey(x, y, zoom):
    quadkey = ""
    for i in range(zoom, 0, -1):
        digit = 0
        mask = 1 << (i - 1)
        if x & mask:
            digit += 1
        if y & mask:
            digit += 2
        quadkey += str(digit)
    return quadkey

# 3. Download single Bing tile
def download_bing_tile_xy(x, y, zoom):
    quadkey = tileXY_to_quadkey(x, y, zoom)
    url = f"https://ecn.t3.tiles.virtualearth.net/tiles/a{quadkey}.jpeg?g=1"

    r = requests.get(url)
    if r.status_code != 200:
        print("Error downloading x=", x, " y=", y)
        return None
    return Image.open(BytesIO(r.content))

# 4. Download NxN tiles around lat/lon & stitch
def download_big_bing_image(lat, lon, zoom, grid_size=3, save_path="big_bing.jpg"):
    cx, cy = latlon_to_tileXY(lat, lon, zoom)

    half = grid_size // 2
    tile_size = 256  # bing tiles are always 256x256

    # Create blank output canvas
    out = Image.new("RGB", (grid_size * tile_size, grid_size * tile_size))

    for done1, dx in enumerate(range(-half, half + 1)):
        for done2, dy in enumerate(range(-half, half + 1)):
            print(f"Downloading tile {done1 * grid_size + done2 + 1} of {grid_size * grid_size}")
            x = cx + dx
            y = cy + dy

            tile_img = download_bing_tile_xy(x, y, zoom)
            if tile_img is None:
                continue

            px = (dx + half) * tile_size
            py = (dy + half) * tile_size

            out.paste(tile_img, (px, py))

    out.save(save_path)
    return out
\end{lstlisting}
\vspace{-1em}
\caption{\textbf{Sky2Ground Python script} used to download satellite images and stitch  aerial-tiles using quadkeys.}
\label{fig:bing_script}
\end{figure*}

\def\suppabstract{
This supplementary document provides additional details 
and extended qualitative results that complement the main Sky2Ground paper.

Section \textcolor{red}{\S\ref{sec:google_studio}} describes our synthetic data generation pipeline 
using Google Earth Studio (GES). We provide expanded discussion of the rendering 
engine, camera parameterization, and the 3D photogrammetric mesh used to generate 
synthetic satellite views with controlled pose, altitude, incidence angle, and 
baseline variations.

Section \textcolor{red}{\S\ref{sec:real_satellite}} details our real satellite image collection process, 
including examples of corresponding synthetic–real pairs. We provide additional 
visualizations illustrating appearance differences, geometric alignment, 
and coverage diversity across multiple scenes. Additionally, we present the complete Python script used for 
collecting high-resolution real satellite imagery from Bing Maps using quadkey-based 
addressing and Web Mercator tiling. 

Section \textcolor{red}{\S\ref{sec:supp_baselines}} provides extended descriptions of the baseline 
methods DUSt3R and MASt3R used in our benchmark, including their architectural 
components, matching mechanisms, and 3D pointmap regression strategies.

Section \textcolor{red}{\S\ref{sec:additional_results}} presents additional experimental results on 
real satellite images, including quantitative comparisons between SkyNet and VGGT 
and an analysis of relative performance drop from synthetic to real imagery.

Finally, we include high-resolution qualitative examples of our dataset in Fig.~\ref{fig:sat_img_1} - \ref{fig:sat_img_3}, including 
paired synthetic–real satellite views and detailed visualizations of ground-truth 
point clouds across multiple Sky2Ground scenes.
}

    \suppabstract

\newcommand{\sref}[1]{Sec.~\ref{#1}}
\setcounter{table}{0}
\renewcommand{\thetable}{A.\arabic{table}}

\section{Synthetic Data Generation via Google Earth Studio}
\label{sec:google_studio}

\indent To supplement real-world aerial imagery, we utilize Google Earth Studio (GES)~\cite{googleearthstudio}, a rendering engine built upon Google’s 3D Earth platform. Unlike standard satellite imagery providers that offer static, nadir-facing 2D tiles~\cite{battersby2014webmercator}, GES operates within a 3D photogrammetric reconstruction of the globe. This allows for the rendering of 2D projections from arbitrary camera poses ($SE(3)$), effectively treating the platform as a simulator for virtual satellite or drone matched-views~\cite{ham2024dragon}.

\subsection{Rendering and Camera Control}

\indent The core utility of GES for our pipeline lies in its precise camera parameterization. The platform grants control over intrinsic and extrinsic parameters, including latitude, longitude, altitude, pitch, yaw, and field of view (FoV). By manipulating these parameters, we can generate datasets that strictly adhere to specific geometric constraints, such as fixed ground sampling distances (GSD) or specific incidence angles, which are often inconsistent in public satellite archives.

The underlying 3D mesh captures complex terrain and urban geometry, including elevation changes and building structures. Consequently, the rendered views inherently account for geometric occlusions and perspective distortions characteristic of low-altitude aerial photography. This provides a higher degree of geometric fidelity compared to warping static 2D satellite tiles (orthophotos) onto digital elevation models (DEMs), where vertical facades are often lost or distorted~\cite{toutin2004geometric,kraus2007photogrammetry}.

\subsection{Dataset Curation and Implementation}
We leverage GES to generate the synthetic component of the Sky2Ground dataset. This approach addresses two limitations of real-world data acquisition:
\begin{itemize}
    \item \textbf{Pose Diversity:} Real satellite imagery is predominantly nadir. GES allows us to systematically sample off-nadir views and varying altitudes, exposing the model to a wider distribution of perspective shifts during training.
    \item \textbf{Controlled Baselines:} We construct multi-view sequences by defining spline-based camera trajectories. This enables the generation of image pairs with known relative transformations, facilitating the evaluation of pose estimation and matching robustness under controlled baseline expansions.
\end{itemize}

The rendering pipeline outputs high-resolution RGB frames along with synchronized metadata files containing the camera state vectors. These synthetic samples serve as a proxy for drone-to-satellite matching scenarios, allowing us to pre-train models on dense geometric variations before fine-tuning on sparser real-world data.

\section{Collection of Real Satellite Images}
\label{sec:real_satellite}

\begin{figure*}[ht!]
    \vspace{-5em}
    \centering
    \setlength{\fboxsep}{4pt} %
    \fbox{%
        \begin{minipage}{\textwidth}
            \centering
            \includegraphics[width=\textwidth]{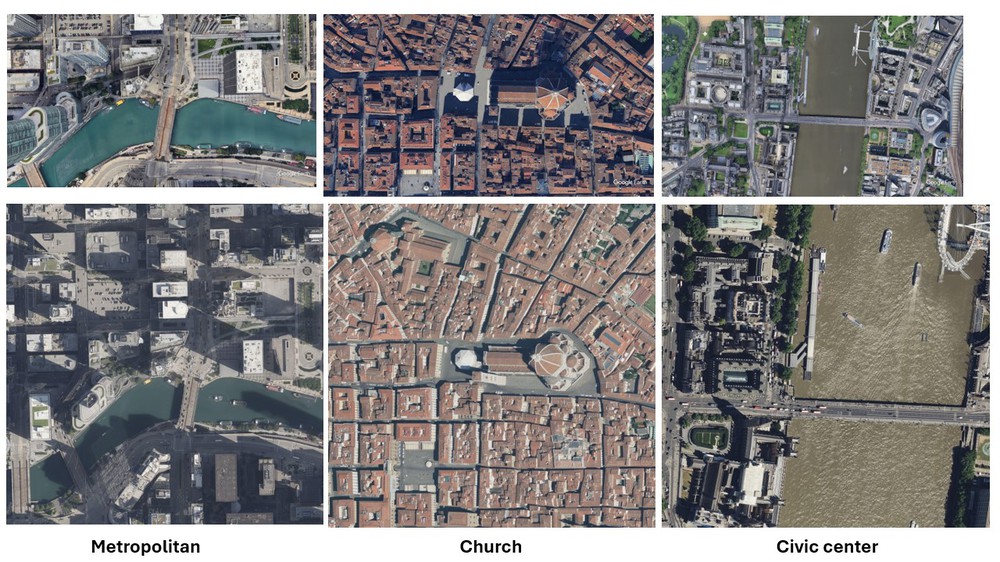}
        \end{minipage}
    }
    \caption{\textbf{(Top row):} synthetic satellite images generated from Google Earth Studio. \textbf{(Bottom row)}: corresponding real satellite images we collected.}
    \vspace{-2em}
    \label{fig:sat_img_1}
\end{figure*}

To support our dataset creation process, we developed a Python script that retrieves high-resolution satellite imagery directly from the Bing Maps server\ref{fig:bing_script} ~\cite{bingmaps}. Due to license issues, we will release our script, and a user can download relevant imagery for personal non-commercial use. 
 This module enables us to programmatically access georeferenced aerial data centered at any geographic coordinate on Earth.
 Because our dataset requires consistent coverage over large areas, we rely on Bing's quadkey tile addressing scheme, together with the Web Mercator grid system, to construct a reproducible and scalable retrieval pipeline.
 In this section, we describe the complete functionality and design of the script in extensive detail.

The script begins by converting a latitude and longitude into what Bing Maps refers to as tile coordinates.
The function \texttt{latlon\_to\_tileXY(lat, lon, zoom)} takes three inputs: a geographic latitude, a geographic longitude, and a zoom level.
Web Mercator, the projection used by most online map systems, divides the world into a grid of square tiles. 
At zoom level $0$, the entire world is represented by a single tile.
At zoom level $1$, the world is divided into a $2 \times 2$ grid. 
At zoom level $2$, it becomes a $4 \times 4$ grid, and so on, doubling along each dimension at every level. 
Therefore, the number of tiles grows rapidly as the zoom level increases, and the function must determine which specific tile contains the given latitude and longitude.

Internally, the conversion process normalizes the longitude so that values ranging from $-180$ degrees to $180$ degrees map smoothly into a horizontal index that spans the total number of tiles available at the chosen zoom level. 
The latitude, on the other hand, must first be passed through the Web Mercator transformation. This transformation converts geographic latitude, which ranges from $-90$ degrees to $+90$ degrees, into a compressed vertical coordinate that accounts for Mercator's nonlinear stretching near the poles. 
The result is a pair of integer indices, referred to as \texttt{x\_tile} and \texttt{y\_tile}, that uniquely identify the horizontal and vertical tile positions.

Once these tile coordinates are known, the next step is to convert them into a quadkey. 
The function \texttt{tileXY\_to\_quadkey(x, y, zoom)} performs this conversion. 
A quadkey is a string composed of the digits $0, 1, 2,$ and $3$. Each character encodes a specific quadrant of a quadtree, which is a hierarchical data structure used extensively in large-scale mapping applications.
At zoom level $1$, there are $4$ possible tiles, and thus $1$ quadkey digit.
At zoom level $2$, each of those tiles is subdivided into $4$ more tiles, requiring a two-character quadkey. 
This pattern continues until the full quadkey, whose length equals the zoom level, is generated. 
The conversion function examines the bits of the x and y tile coordinates at each level of the hierarchy.
Depending on the bit pattern, it appends the corresponding digit to the quadkey string. 
For example, a zero digit indicates the upper-left quadrant, while higher digits refer to other combinations of horizontal and vertical offsets.
By the end of the loop, the quadkey is a compact and unambiguous identifier that Bing Maps uses to index satellite tiles.
\begin{figure*}[ht!]

    \centering
    \setlength{\fboxsep}{4pt} %
    \fbox{%
        \begin{minipage}{\textwidth}
            \centering
            \includegraphics[width=\textwidth]{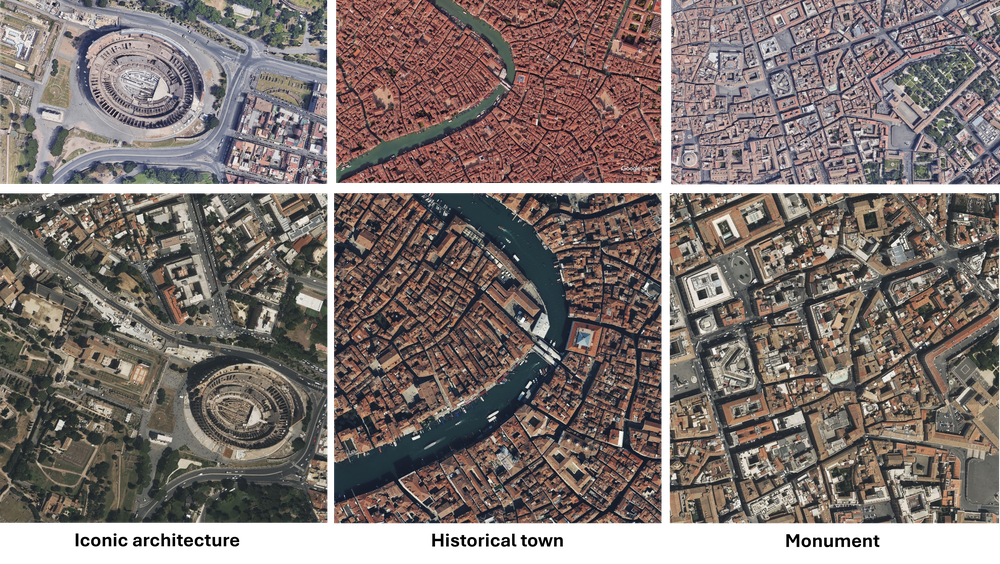}
        \end{minipage}
    }
    \caption{\textbf{(Top row):} synthetic satellite images generated from Google Earth Studio. \textbf{(Bottom row)}: corresponding real satellite images we collected.}
    \label{fig:sat_img_2}
\end{figure*}
The third component of the script, \texttt{download\_bing\_tile\_xy(x, y, zoom)}, is responsible for retrieving a single aerial tile from the Bing Maps server. 
Using the quadkey computed in the previous step, the function assembles a complete URL that follows Bing's documented tile endpoint format. 
The URL includes the quadkey as part of the file path, allowing the server to respond with the exact tile requested.
The function then performs an HTTP request and checks the returned status code. 
If the server responds successfully, the tile is decoded as a JPEG image and returned as a standard Python Imaging Library object. 
If there is an error, such as an invalid tile index or a network failure, the function reports the issue and returns a null value. This design choice allows the system to handle missing tiles gracefully without interrupting larger download tasks.

While a single tile may suffice for many applications, our dataset construction pipeline often requires a much larger spatial footprint. To achieve this, the final component of the module, \texttt{download\_big\_bing\_image(lat, lon, zoom, grid\_size)}, stitches together an entire grid of tiles around a central geographic coordinate. The user specifies a grid size, such as $3 \times 3$, $5 \times 5$, or $10 \times 10$. 
The function first determines the tile coordinates of the center position using the initial conversion function. 
It then computes a range of offsets centered around this tile. For example, if the grid size is $5$, the offsets range from $-2$ to $+2$ along both the horizontal and vertical directions.

The function initializes an empty output canvas whose width and height depend on the number of tiles requested. Since each tile is $256 \times 256$ pixels, a $5 \times5$ grid yields an output image of $1280$ pixels on each side. The function then begins iterating over all tile offsets. For each offset pair, it computes the corresponding tile index, downloads the tile using the previously described component, and pastes the resulting image into the correct location on the output canvas. This continues until all tiles have been downloaded and positioned. Any tile that fails to download simply leaves an empty or unchanged region in the composite image, which is acceptable for our use case.

After all tiles have been processed, the final composite image is saved to disk in JPEG format. This large image provides a continuous satellite view centered at the specified coordinates, covering an area that scales directly with the chosen zoom level and grid size. 
Such images are instrumental in our dataset pipeline, as they offer consistent spatial coverage, controlled resolution, and reliable alignment with real-world coordinates. Unlike screenshots or manually captured satellite images, the quadkey-based retrieval ensures deterministic and repeatable results.

\begin{figure*}[ht!]

    \centering
    \setlength{\fboxsep}{4pt} %
    \fbox{%
        \begin{minipage}{\textwidth}
            \centering
            \includegraphics[width=\textwidth]{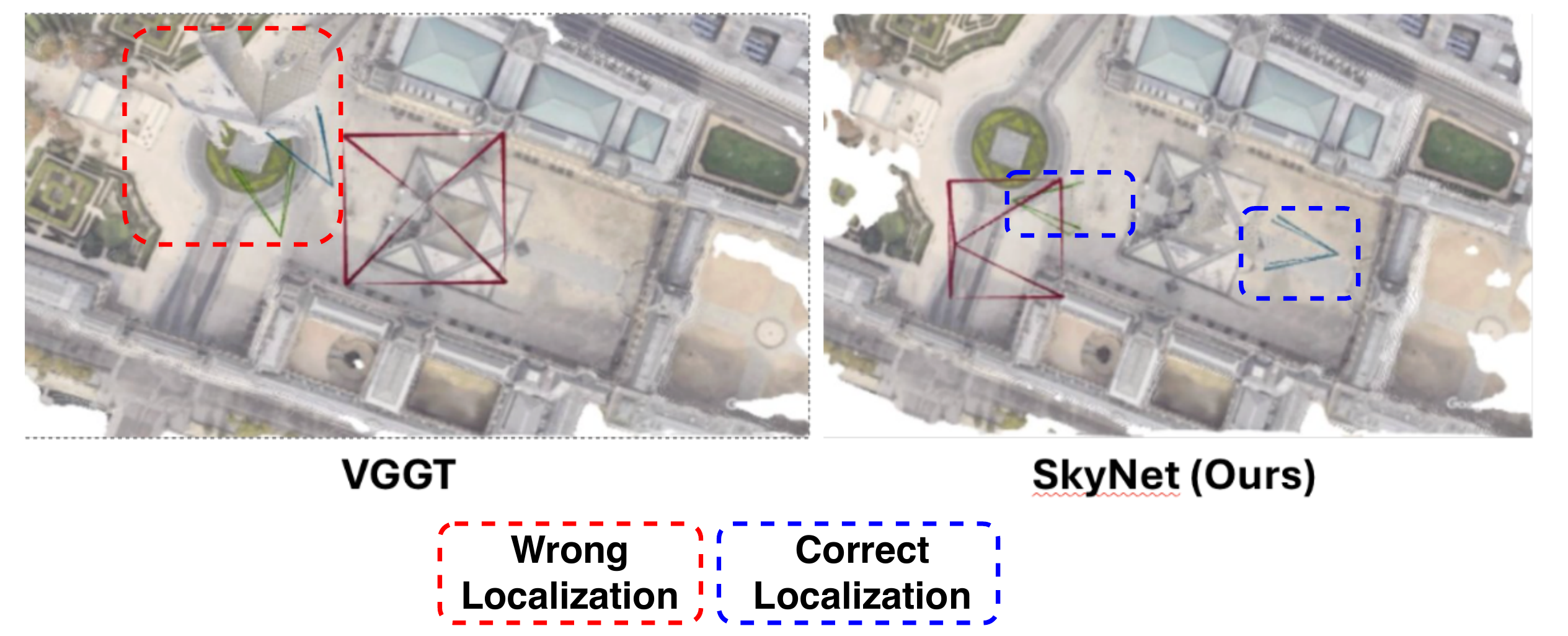}
        \end{minipage}
    }
    \caption{\textbf{Qualitative results of SkyNet vs VGGT:} Given one satellite image, and two ground views, VGGT fails to localize the pointmaps properly, for eg, the pyramid is formed at wrong location \textcolor{red}{(marked in red)}, in contrast, SkyNet correctly localizes the ground views \textcolor{blue}{(marked in blue)}, even when the views are non-overlapping. }
    \label{fig:sat_img_2}
\end{figure*}

\section{Baseline Methods in Sky2ground Benchmark}
\label{sec:supp_baselines}

We benchmark Sky2Ground against state-of-the-art feed-forward geometry models that unify dense correspondence, depth, and pose estimation within a single transformer backbone. Specifically, we utilize DUSt3R and its matching-focused extension, MASt3R.

\subsection{DUSt3R}

DUSt3R\cite{dust3r_cvpr24} reformulates two-view reconstruction as a dense pointmap regression task. Rather than enforcing a strict projective camera model with known intrinsics, the model predicts a per-pixel 3D point ($X, Y, Z$) for each view in a canonical, camera-agnostic coordinate system. The network also outputs confidence scores to indicate geometric reliability. By eschewing explicit camera modeling, DUSt3R learns to generate pointmaps where mutual consistency implicitly encodes the underlying depth and viewpoint changes.

The architecture employs a symmetric two-branch transformer encoder based on pretrained CroCo ViT features~\cite{croco,croco_v2}. A transformer decoder performs dense cross-view reasoning via cross-attention, allowing tokens from one view to attend to all tokens from the other. This produces fused representations that capture soft correspondences across the entire image plane. The decoder heads then regress the dense pointmaps and confidences. This global reasoning enables robustness to wide baselines and partial overlaps, as the model does not rely on a brittle set of sparse feature matches.

To resolve the coordinate ambiguity between views, DUSt3R applies a \textbf{global alignment} procedure. For a given pair, the model estimates the similarity transform (rotation, translation, and scale) that best aligns the two predicted point clouds using a weighted Procrustes fit, with weights derived from the predicted confidence scores. For multi-view setups, this is extended to a global optimization that aligns all pairwise maps into a common world frame. While this approach offers significant stability under extreme viewpoint changes, the pixel-level precision of the implicit matches can be lower than that of specialist feature matchers.

\subsection{MASt3R}

MASt3R\cite{mast3r_arxiv24} extends the DUSt3R architecture to support \emph{matching-first} localization while retaining the feed-forward geometric robustness of the original model. The authors address the limitation that DUSt3R, while effective for global alignment, often yields correspondences that are too smooth for high-precision visual localization.

MASt3R retains the pointmap regression head of DUSt3R but introduces a second \textbf{dense local descriptor head}. This head predicts per-pixel feature maps and is trained with an InfoNCE-style contrastive objective, forcing descriptors to be discriminative and locally precise. This results in a hybrid output: dense 3D pointmaps for geometric grounding and distinct feature descriptors for fine-grained matching.

During inference, matches are extracted via a \textbf{coarse-to-fine reciprocal matching scheme}. Initial correspondences are identified at a low resolution using the descriptor maps and subsequently refined at higher resolutions. The pipeline enforces reciprocity—requiring matches to be mutual under nearest-neighbor lookup—to suppress outliers effectively. In our Sky2Ground setting, this distinction is critical: while both methods inherit robustness to wide baselines, MASt3R consistently outperforms DUSt3R when accurate pixel-level correspondences are a prerequisite for pose estimation.
Representative examples of these collected real satellite images are shown in Fig.~\ref{fig:sat_img_1} - \ref{fig:sat_img_3} of the supplementary material.

\section{Additional Results}
\label{sec:additional_results}

\begin{table}[ht!]
\centering
\scriptsize
\setlength{\tabcolsep}{0.8pt}
\caption{\textbf{Evaluation on real satellite-images, across Ground+Satellite setup}
RRA@5 / RTA@5 denote rotation and translation accuracy within 5° / 5 m. SkyNet has lower relative-drop in performance than the VGGT-baseline.}
\begin{tabular}{lcccccccccc}
\toprule
\textbf{Method} & \textbf{Data} &
\multicolumn{3}{c}{\textbf{Ground}} &
\multicolumn{3}{c}{\textbf{Satellite}} &
\multicolumn{2}{c}{\textbf{Avg}} \\
\cmidrule(lr){3-5} \cmidrule(lr){6-8} \cmidrule(lr){9-10}
& & RRA & RTA & Avg & RRA & RTA & Avg & Overall Avg & Drop \\
\midrule
\textbf{Synth-Sat}\\
VGGT\cite{wang2025vggt} & Sky2Ground & 34.5 & 40.2 & 37.4 & 83.3 & 33.3 & 58.3 & 47.8 & -- \\
\rowcolor{gray!20}
\textbf{SkyNet} & \textbf{Sky2Ground} & 71.3 & 64.9 & 68.1 & 88.0 & 36.3 & 62.2 & \textbf{65.1} & -- \\
\hline
\textbf{Real-Sat} \\
VGGT & Sky2Ground & 30.1 & 37.8 & 33.9 & 62.4 & 25.6 & 44.0 & 38.9 & \textcolor{mypink}{$-18.6\%$} \\
\rowcolor{gray!20}
\textbf{SkyNet} & \textbf{Sky2Ground} & 69.0 & 62.7 & 65.8 & 84.1 & 33.2 & 58.6 & \textbf{62.2} & \textcolor{mypink}{$-4.0\%$} \\
\bottomrule
\end{tabular}
\label{tab:gs_real}
\end{table}

In Table \ref{tab:gs_real}, we compare our method (SkyNet) against the VGGT baseline for localization within the ground-satellite setup using real satellite images~\cite{wang2025vggt}. 
We observe that performance across all models drops when transitioning from synthetic to real imagery, underscoring the inherent domain gap challenges.

However, SkyNet demonstrates significantly higher robustness: our method experiences a marginal relative drop of only $4.0\%$ in overall average performance, whereas the VGGT baseline drops by $18.6\%$. This smaller performance decay highlights SkyNet's superior efficacy and its ability to generalize more effectively to real-world satellite data.

\section{License and Usage Statement}

The Sky2Ground dataset will be released under a Creative Commons 
Attribution–NonCommercial (CC BY-NC) license. The dataset is intended 
solely for non-commercial research, educational, and scientific use. 
Users are permitted to copy, redistribute, and adapt the dataset for 
scholarly work provided that appropriate attribution is given to the 
authors and commercial usage of any kind is strictly prohibited.

The real satellite imagery used in this project is sourced from the 
Bing Maps imagery service. As we do not own the rights to redistribute 
these images, they are not included directly with the dataset release. 
Instead, we provide our full data collection script, which uses Bing’s 
public quadkey tile interface to allow users to download imagery 
themselves strictly for personal, non-commercial use in accordance 
with Microsoft's licensing terms.

In addition to the dataset, we will publicly release our evaluation 
benchmark, protocol, and baseline model implementations on GitHub. This release helps
support reproducible research and enable the community to build upon 
our benchmark.

\newpage

\begin{figure*}[ht]
    \centering
    \setlength{\fboxsep}{4pt} %
    \fbox{%
        \begin{minipage}{\textwidth}
            \centering
\includegraphics[width=\textwidth,height=0.8\textheight]{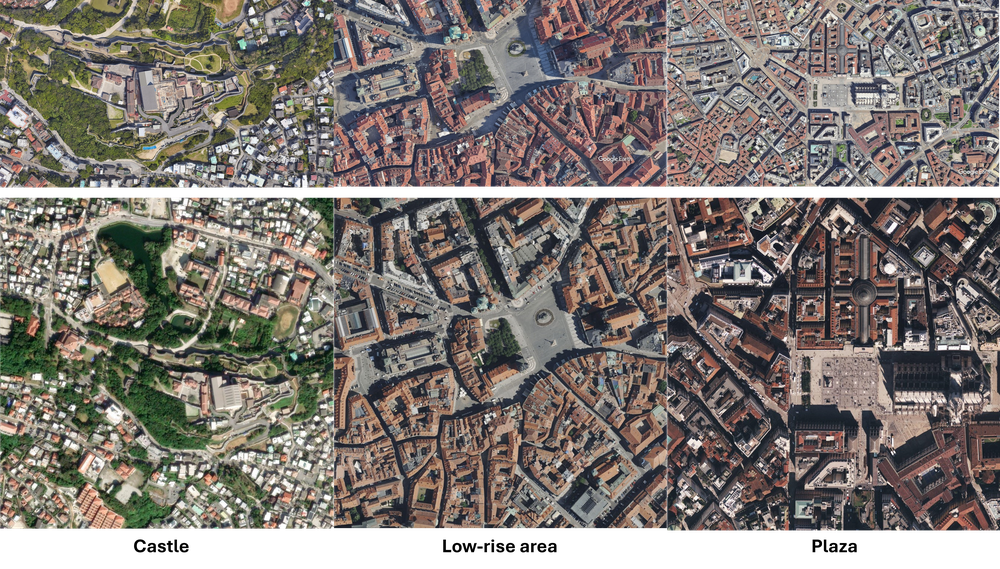}
        \end{minipage}
    }
    \caption{\textbf{(Top row):} synthetic satellite images generated from Google Earth Studio. \textbf{(Bottom row)}: corresponding real satellite images collected from aerial sources.}
    \label{fig:sat_img_3}
\end{figure*}

\begin{figure*}[ht]
    \centering
    \fbox{%
        \begin{minipage}{\textwidth}
            \centering
            \includegraphics[width=\textwidth,height=0.9\textheight]{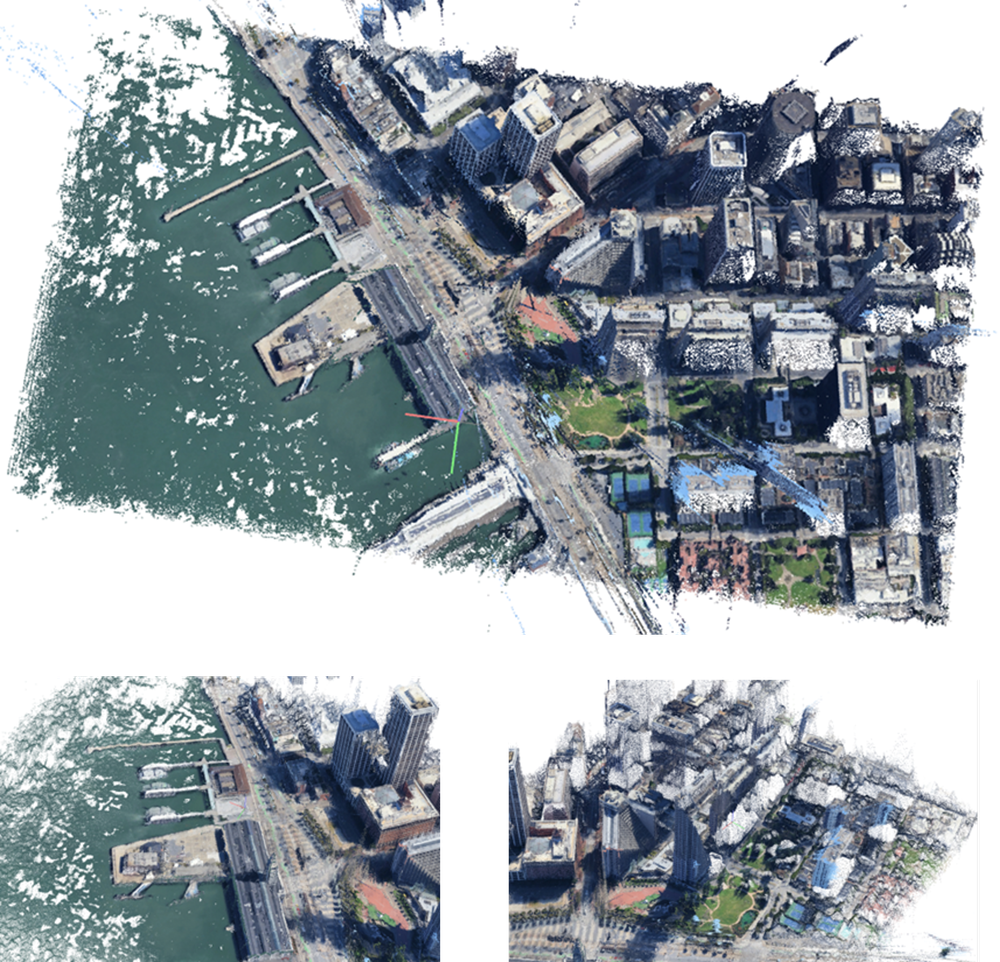}
        \end{minipage}
    }
    \caption{\textbf{Ferry Building}: Visualization of the ground-truth point cloud on our Sky2Ground dataset. The top row shows the reconstructed scene from a satellite view perspective, and the bottom row presents two aerial view renderings for structural details and geometry.}
    \label{fig:qual_3}
\end{figure*}

\newpage

    \begin{figure*}[ht]
    \centering
    \setlength{\fboxsep}{4pt} %
    \fbox{%
        \begin{minipage}{\textwidth}
            \centering
\includegraphics[width=\textwidth,height=0.9\textheight]{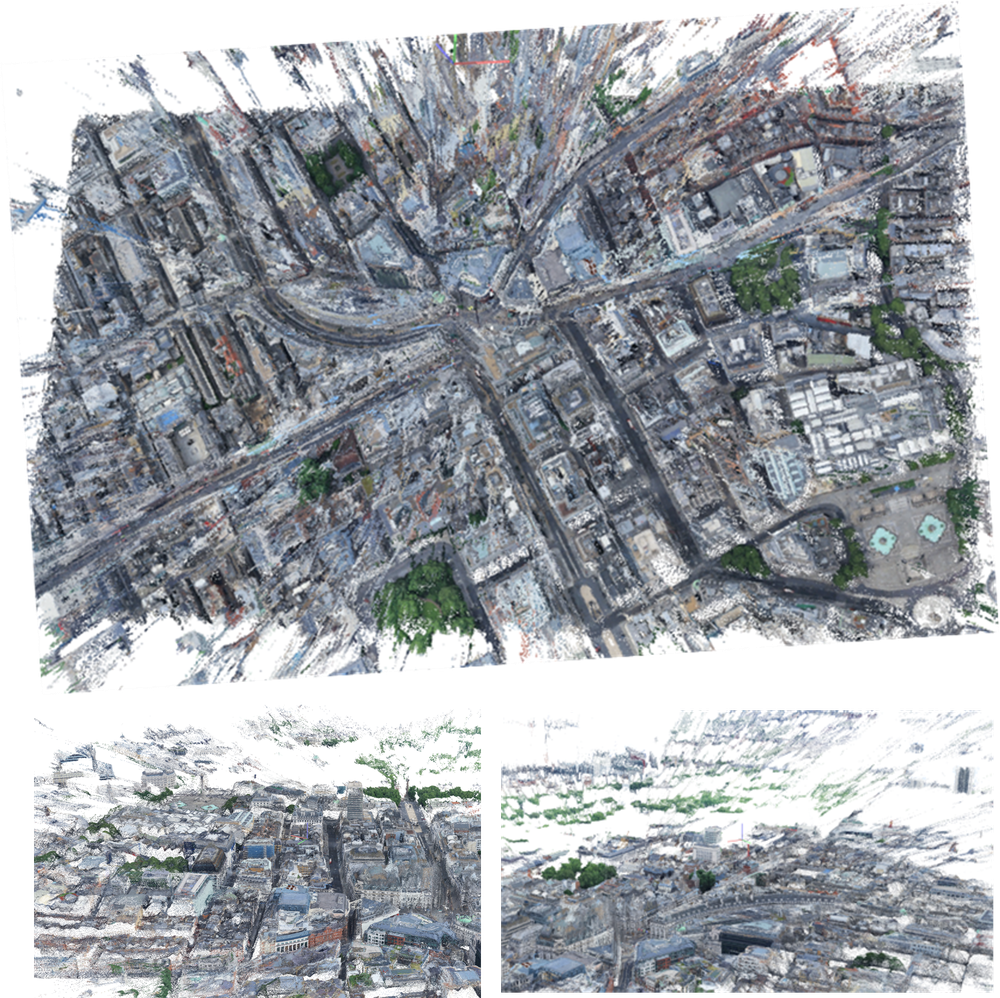}
        \end{minipage}
    }
    \caption{\textbf{Trafalgar Square}: Visualization of the ground-truth point clouds on our Sky2Ground dataset. The top row shows the reconstructed scene from a satellite view perspective, and the bottom row presents two aerial view renderings for structural details and geometry.}
    \label{fig:qual_2}
\end{figure*}

\newpage

\newpage

\begin{figure*}[ht]
    \centering
    \setlength{\fboxsep}{4pt} %
    \fbox{%
        \begin{minipage}{\textwidth}
            \centering
            \includegraphics[width=\textwidth,height=0.9\textheight]{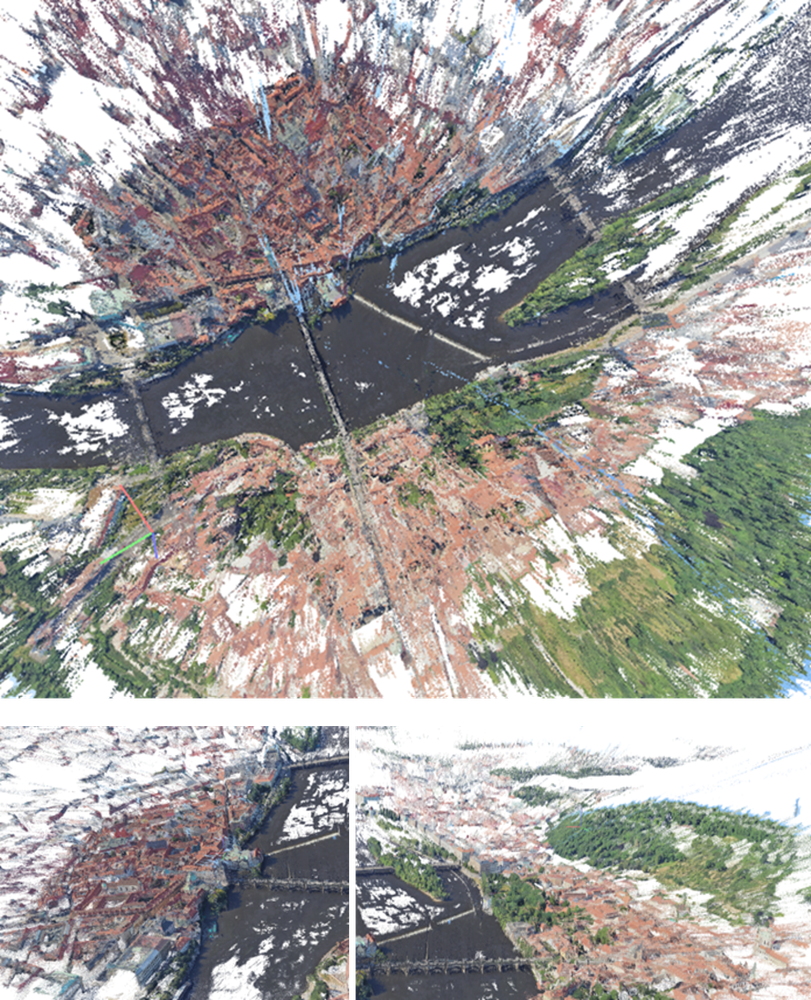}
        \end{minipage}
    }
    \caption{\textbf{Charles Bridge}: Visualization of the ground-truth point cloud on our Sky2Ground dataset. The top row shows the reconstructed scene from a satellite view perspective, and the bottom row presents two aerial view renderings for structural details and geometry.}
    \label{fig:qual_4}
\end{figure*}
\newpage

\begin{figure*}[ht]
    \centering
    \setlength{\fboxsep}{4pt}
    \fbox{%
        \begin{minipage}{\textwidth}
            \centering
            \includegraphics[width=\textwidth,height=0.9\textheight]{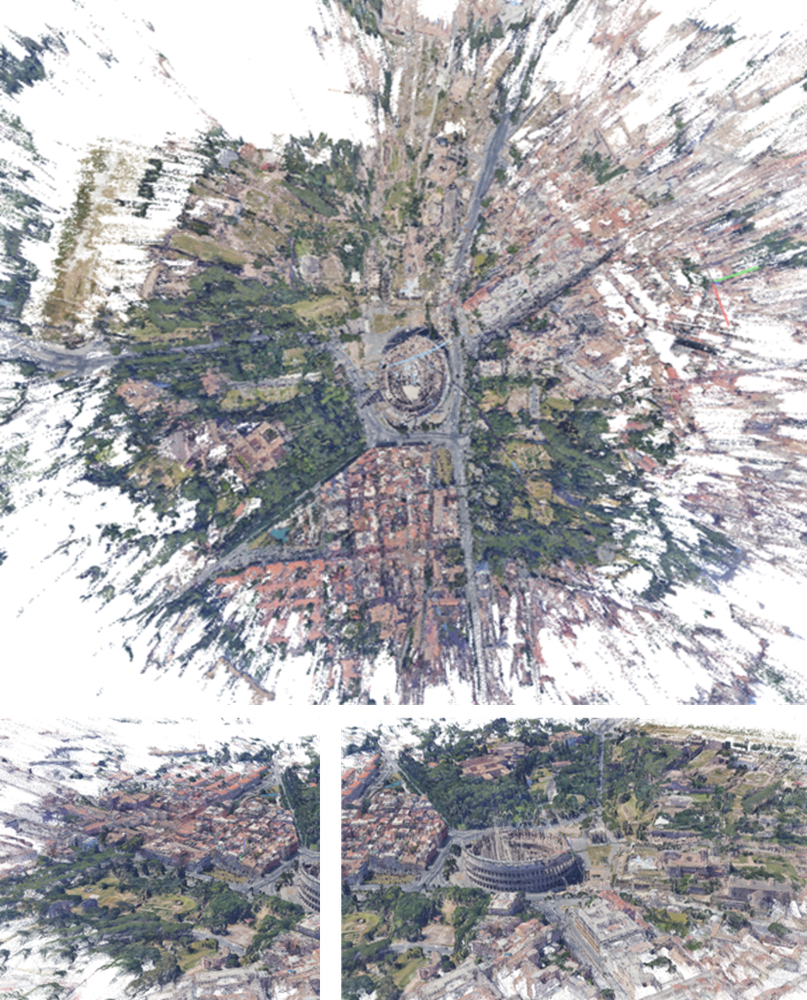}
        \end{minipage}
    }
    \caption{\textbf{Colosseum}: Visualization of the ground-truth point cloud on our Sky2Ground dataset. The top row shows the reconstructed scene from a satellite view perspective, and the bottom row presents two aerial view renderings for structural details and geometry.}
    \label{fig:qual_5}
\end{figure*}
\newpage

\begin{figure*}[ht]
    \centering
    \setlength{\fboxsep}{4pt}
    \fbox{%
        \begin{minipage}{\textwidth}
            \centering
            \includegraphics[width=\textwidth,height=0.9\textheight]{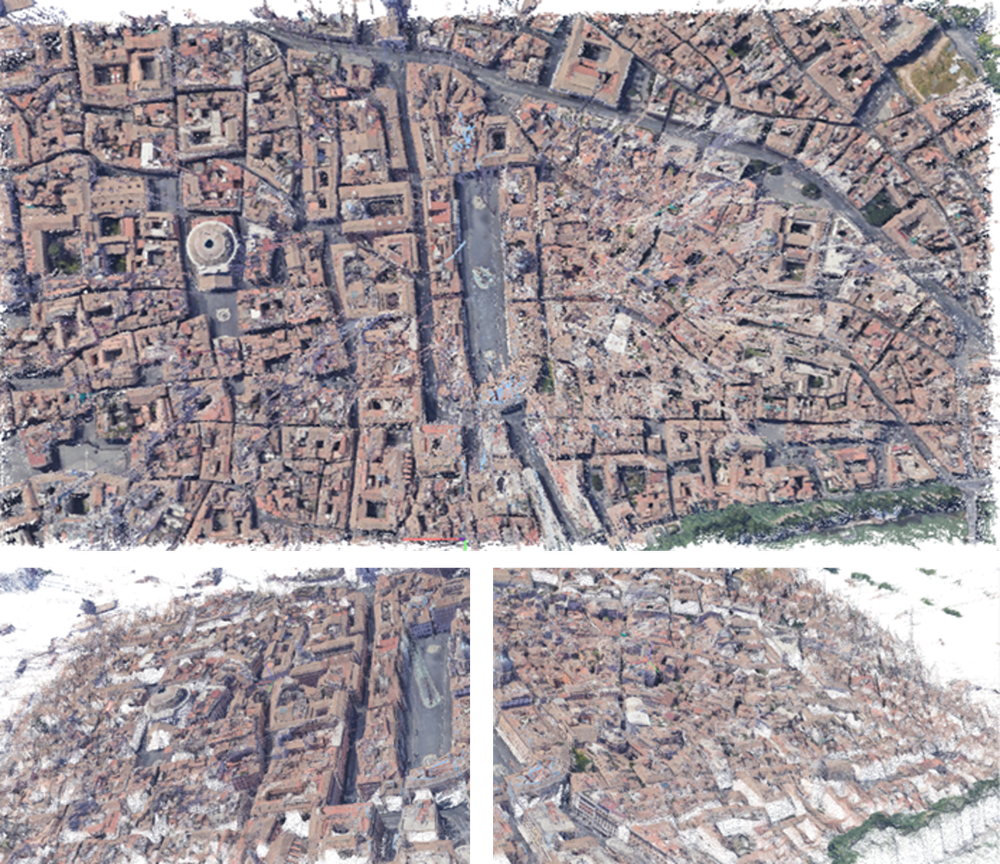}
        \end{minipage}
    }
    \caption{\textbf{Piazza Navona}:Visualization of the ground-truth point cloud on our Sky2Ground dataset. The top row shows the reconstructed scene from a satellite view perspective, and the bottom row presents two aerial view renderings for structural details and geometry.}
    \label{fig:qual_6}
\end{figure*}

\cleardoublepage

\end{document}